\documentclass[10pt,twocolumn,letterpaper]{article}

\usepackage[pagenumbers]{cvpr}

\usepackage{colortbl}

\usepackage{tabularx}

\usepackage{stfloats}
\usepackage{multirow}

\usepackage{algorithm}
\usepackage{algpseudocode}
\usepackage{xcolor}
\usepackage{tabularx}

\definecolor{cvprblue}{rgb}{0.21,0.49,0.74}
\usepackage[pagebackref,breaklinks,colorlinks,allcolors=cvprblue]{hyperref}

\title{Are We Rewarding the Wrong Module? \\ Module-Specific Policy Optimization for Multimodal Reinforcement Learning}

\author{
HuiSeong Cho\thanks{Equal contribution. Author ordering determined by coin flip.}
\qquad
Jun Yeong Park\footnotemark[1]
\qquad
Yu Rang Park\thanks{Corresponding author.}
\\
Yonsei University, Seoul, Korea
\\
{\tt\small 2026311565@yonsei.ac.kr pjk000011@naver.com yurangpark@yuhs.ac}
}

\begin{document}
\maketitle
\begin{abstract}
Reinforcement learning with verifiable rewards (RLVR) based on Group Relative Policy Optimization (GRPO) has been widely adopted to enhance the multimodal reasoning of multimodal large language models (MLLMs). However, existing methods share a common limitation of Module-Level Credit Entanglement, in which the vision encoder, projector, and LLM modules are jointly optimized using the same advantage signals even when perceptual correctness and final-answer correctness differ. To address this limitation, we propose \textbf{M}odule-\textbf{S}pecific \textbf{P}olicy \textbf{O}ptimization (\textbf{MSPO}), a framework that separates perceptual and outcome-based learning signals across model modules. Specifically, MSPO uses responses to perception checklists comprising auxiliary questions about visual information to derive an advantage for training the vision encoder. From these responses, MSPO computes the Confidence-aware Perception Reward (CaPR), using a perception sign obtained by thresholding checklist accuracy to determine reward direction and posterior confidence from a Beta distribution over latent perception accuracy to scale reward magnitude. We then derive the vision advantage through group-relative normalization of CaPR and scale it using main-task success rates. This advantage guides updates to the vision encoder, while the projector and LLM retain outcome-based GRPO updates. Extensive experiments on seven mathematical and general reasoning benchmarks demonstrate that MSPO outperforms existing RL-tuned models on average at both the 3B and 7B scales.
\end{abstract}

\vspace{-2.5mm}
\section{Introduction}

Multimodal reasoning is a fundamental capability of multimodal large language models (MLLMs) for addressing complex vision-language tasks, such as geometric problem solving and chart interpretation~\cite{srpo,r1_onevision,wang2025multimodal,hintgrpo,insight_03}. Recently, to enhance the multimodal reasoning capabilities of MLLMs, reinforcement learning with verifiable rewards (RLVR)~\cite{RLVR} methods based on Group Relative Policy Optimization (GRPO)~\cite{deepseekmath} have been widely adopted.

\begin{figure}[tb]
    \centering
    \includegraphics[width=1.0\linewidth]{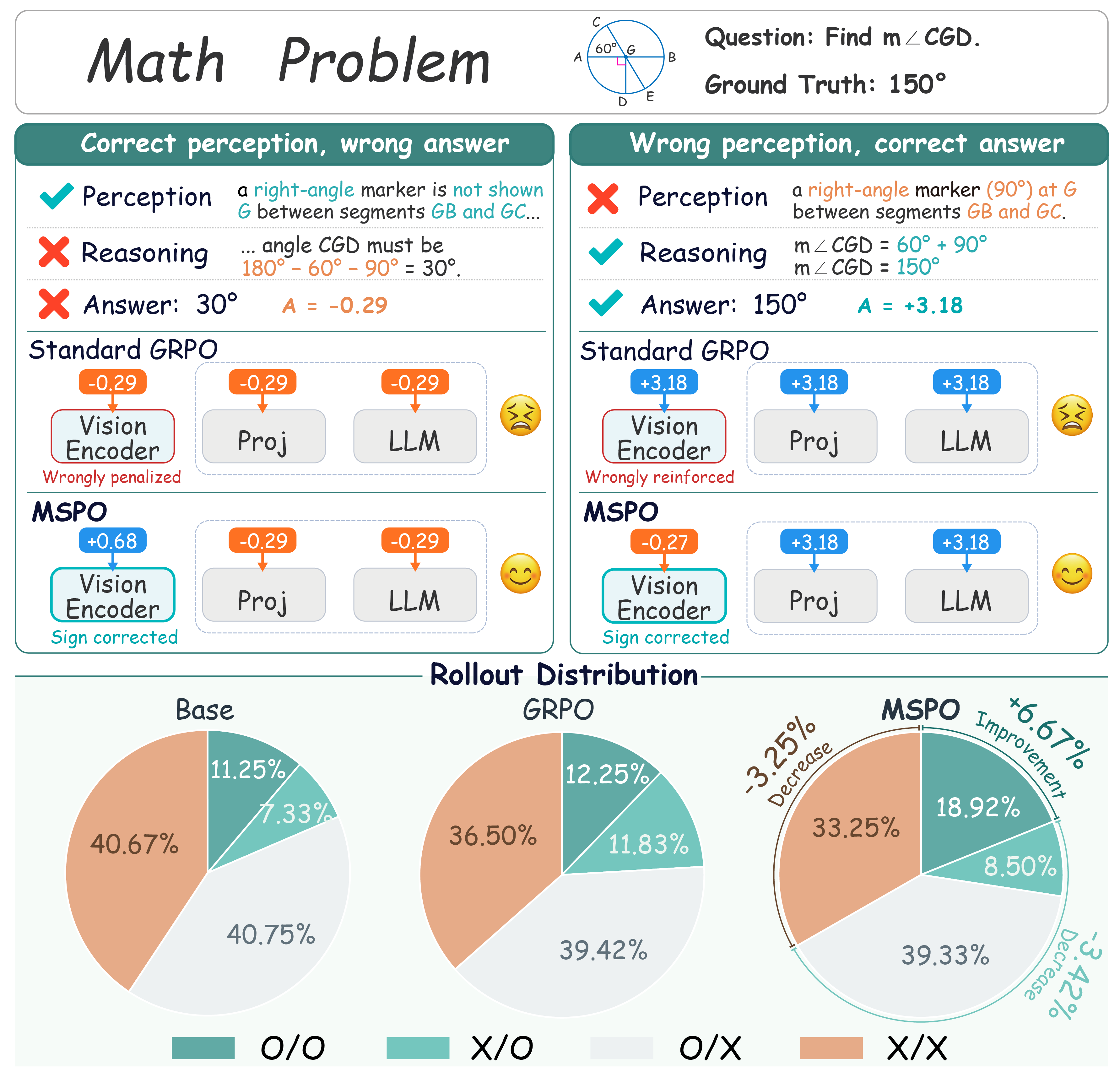}
    \caption{\textbf{Top:} Module-Level Credit Entanglement in GRPO can penalize correct perception or reinforce faulty perception through shared outcome-based advantages. Our MSPO assigns perception-based advantages to the vision encoder while retaining outcome-based advantages for the projector and LLM. \textbf{Bottom:} Rollout distributions on the Geo3K validation set (perception/answer; O: correct, X: incorrect). Compared with GRPO, MSPO reduces overall conflicts (X/O and O/X) and X/X cases while increasing O/O.
}
\label{intro_figure}
\vspace{-1.0em}
\end{figure}

However, conventional RLVR frameworks~\cite{deepseekmath,dapo,visual_rft,dr_grpo} derive advantages from final-answer correctness, without identifying whether errors originate in visual perception or logical reasoning. To overcome this limitation, PEARL~\cite{pearl} introduces a perception checklist to evaluate perceptual correctness and uses the resulting perceptual rewards to optimize an auxiliary perception objective and guide reasoning updates. MoCA~\cite{moca} evaluates visual descriptions with a text-only reasoner to distinguish perceptual from reasoning errors and assign differentiated advantages to perception and reasoning tokens.

Despite these advances in response- or token-level credit assignment, existing RLVR frameworks do not explicitly separate learning signals for the vision encoder, projector, and LLM modules. Even when perception and reasoning tokens receive different advantages, the associated gradients can still update the same model modules. When these modules are jointly optimized, applying the same advantage across modules entangles their credit assignment without explicitly accounting for their respective contributions to the outcome, a phenomenon we term Module-Level Credit Entanglement. As shown in \cref{intro_figure} (Top), the vision encoder may provide sufficient visual information, yet still receive negative credit when the LLM produces an incorrect answer due to a reasoning error. Conversely, faulty perception can receive positive credit when the final answer is correct.

To address this limitation, we propose \textbf{M}odule-\textbf{S}pecific \textbf{P}olicy \textbf{O}ptimization (\textbf{MSPO}), which assigns learning signals according to each module's role. MSPO first uses separately generated checklist responses to assess visual perception without relying on final-answer correctness. To capture perceptual correctness and uncertainty in checklist assessments, we introduce the Confidence-aware Perception Reward (CaPR), whose direction is determined by a perception sign obtained by thresholding checklist accuracy. Posterior confidence scales CaPR's magnitude based on how strongly a Beta posterior over latent perception accuracy supports either side of the threshold. We then derive the Vision Advantage through group-relative normalization of CaPR and apply it exclusively to the vision encoder, so that perceptual correctness guides its learning. Meanwhile, the projector and LLM retain outcome-based GRPO updates. As shown in \cref{intro_figure} (Bottom), MSPO reduces perception--answer conflicts and responses with both incorrect perception and incorrect answers while increasing responses with both correct perception and correct answers compared with GRPO. By aligning optimization with each module's role, MSPO alleviates Module-Level Credit Entanglement and outperforms existing RL-tuned models on average across seven mathematical and general reasoning benchmarks at both the 3B and 7B scales.

Our key contributions are summarized as follows:

\begin{enumerate}
    \item \textit{\textbf{Defining Module-Level Credit Entanglement.}} We are the first to define Module-Level Credit Entanglement, a limitation in multimodal RL arising from shared advantage signals across distinct modules.
    \item \textit{\textbf{Introducing Module-Specific Policy Optimization (MSPO).}} We propose MSPO, a novel framework addressing Module-Level Credit Entanglement by guiding the vision encoder module with advantages from the CaPR while retaining outcome-based GRPO updates for the projector and LLM modules.
    \item \textit{\textbf{Superior Performance across Benchmarks and Model Scales.}} MSPO outperforms existing RL-tuned models on average across seven mathematical and general reasoning benchmarks at both the 3B and 7B scales.
\end{enumerate}
\section{Related work}
\subsection{Multimodal large language models}
Multimodal large language models (MLLMs) have made remarkable progress in various vision-language tasks, including visual question answering (VQA) and complex reasoning~\cite{openai_gpt5,gemini2_5,qwen3,internvl3_5,llava,llava-ov}. MLLMs commonly connect a pretrained vision encoder to an LLM through a learnable projector, using alignment training and instruction tuning~\cite{llava,minigpt,vila,qwen2_5}. The vision encoder extracts visual features, the projector maps them into the LLM's input space, and the LLM uses these representations for reasoning and response generation. Building on this architecture, LLaVA-1.5~\cite{llava1_5}, LLaVA-OneVision~\cite{llava-ov}, Qwen-VL~\cite{qwen,qwen2,qwen2_5,qwen3}, and InternVL~\cite{internvl,internvl3,internvl3_5} further advance multimodal capabilities by scaling up training data and increasing its diversity.

\raggedbottom
\subsection{Multimodal large language model reasoning}
With the advancement of MLLMs, recent studies have drawn inspiration from progress in LLM reasoning and adopted GRPO-based reinforcement learning to enhance multimodal reasoning capabilities. MM-Eureka~\cite{mmeureka} applies rule-based reinforcement learning to multimodal mathematical reasoning, while R1-VL~\cite{r1_vl} introduces dense step-wise reasoning rewards. PAPO~\cite{papo} and VPPO~\cite{vppo} compare model predictions under original and perturbed visual inputs to shape learning signals, and AT-RL~\cite{at_rl} uses cross-modal connectivity to concentrate credit on tokens associated with visual evidence. To explicitly assess perceptual correctness, Perception-R1~\cite{perception_r1} introduces rewards based on consistency with visual annotations, while PEARL~\cite{pearl} evaluates task-relevant perception through checklists. MoCA~\cite{moca} distinguishes perceptual failures from reasoning failures and assigns differentiated credit to the corresponding stages. However, these approaches do not explicitly decouple credit assignment across modules. MSPO addresses this module-level credit entanglement by assigning module-specific advantages.

\section{Preliminary}
\subsection{Group Relative Policy Optimization (GRPO)}

\textit{Group Relative Policy Optimization (GRPO)}~\cite{deepseekmath} is widely used in RLVR. Building on Proximal Policy Optimization (PPO)~\cite{ppo}, it estimates advantages through within-group reward comparisons without learning a separate value function. Given a query $q$, a group of $G$ responses $\{o_i\}_{i=1}^{G}$ is sampled from the old policy $\pi_{\theta_{\mathrm{old}}}$.
The $i$-th response $o_i$ receives an outcome-level reward $R_i$, which is normalized within the group to obtain the advantage:

{\small
\begin{equation}
\label{eq:GRPO_advantage}
\hat{A}_i =
\frac{
R_i - \mathrm{mean}\left(\{R_k\}_{k=1}^{G}\right)
}{
\mathrm{std}\left(\{R_k\}_{k=1}^{G}\right)
}.
\end{equation}
}
The policy is updated by maximizing the following clipped objective:
{\small
\begin{multline}
\label{eq:GRPO_objective}
\displaystyle \mathcal{J}_{\mathrm{GRPO}}(\theta) = \mathrm{E}_{q \sim \mathcal{D}, \{o_i\}_{i=1}^G \sim \pi_{\theta_{\mathrm{old}}}(\cdot|q)} \Bigg[ \frac{1}{G} \sum_{i=1}^G \frac{1}{|o_i|} \sum_{t=1}^{|o_i|}\bigg( \\
\min \Big( \rho_{i,t} \hat{A}_i,
\mathrm{clip}(\rho_{i,t}, 1-\epsilon_{\mathrm{clip}}, 1+\epsilon_{\mathrm{clip}}) \hat{A}_i \Big) \\
- \beta D_{\mathrm{KL}}(\pi_\theta \| \pi_{\mathrm{ref}}) \bigg) \Bigg]
\end{multline}
}
where $\rho_{i,t}(\theta)
=
\frac{
\pi_{\theta}(o_{i,t}\mid q,o_{i,<t})
}{
\pi_{\theta_{\mathrm{old}}}(o_{i,t}\mid q,o_{i,<t})
}$
denotes the probability ratio, and $\beta$ controls the strength of the KL penalty.
\begin{figure*}[t]
    \centering
    \vspace{-3.0mm}
    \includegraphics[width=\textwidth]{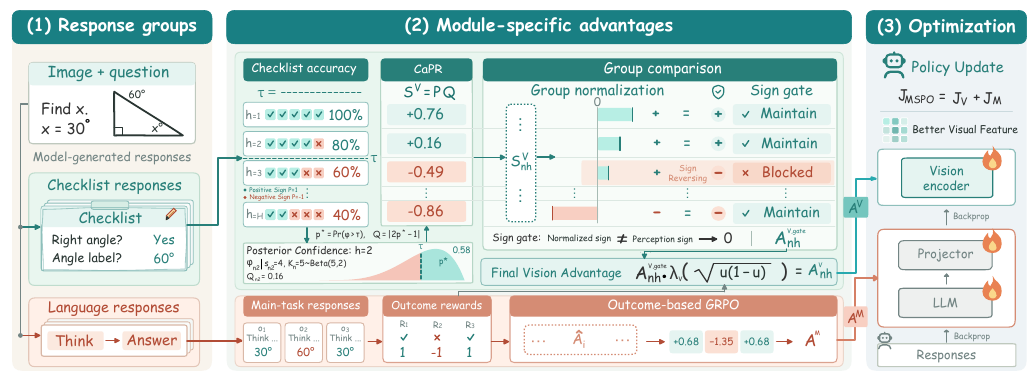}
    \vspace{-4mm}
    \caption{\textbf{The framework of MSPO.} (1) The model separately generates checklist and main-task responses.
(2) The Confidence-aware Perception Reward (CaPR) combines the Perception Sign with posterior confidence. Group normalization, sign gating, and main-task success-rate modulation produce the vision advantage, while final-answer rewards yield the outcome-based GRPO advantage.
(3) These advantages guide updates to the vision encoder and the projector and LLM, respectively, in a single optimization step.}
    \label{fig:mspo_overview}
    \vspace{-1.7mm}
\end{figure*}

\raggedbottom

\section{Method}
\label{sec:method}

\subsection{Overview}
\label{sec:method_overview}\cref{fig:mspo_overview} illustrates the overall framework of Module-Specific Policy Optimization (MSPO), which addresses Module-Level Credit Entanglement through module-specific advantage signals. We first construct perception checklists covering task-relevant facts directly observable in each image. During training, the model separately generates responses to the original task questions and each checklist question. We derive the Confidence-aware Perception Reward (CaPR) from checklist correctness and posterior confidence and use it to construct the Vision Advantage. We then adjust the magnitude of the vision advantage according to the success rate of the corresponding main-task rollout group. The outcome-based advantage is computed from final-answer correctness using standard GRPO. Finally, we update the vision encoder module with the vision advantage and the projector and LLM modules with the outcome-based advantage in a single optimization step.

\subsection{Perception checklist construction}
\label{sec:visual_checklist}

We use perception checklists~\cite{pearl} consisting of auxiliary question--answer pairs about visual information in the original problem, such as \textit{``What length is marked on $AB$?''} paired with \textit{``10''}. For problem $n$, we denote its checklist as $\mathcal{C}_n=\{(c_{nj},b_{nj})\}_{j=1}^{K_n}$, where $c_{nj}$ is a checklist question, $b_{nj}$ is its ground-truth answer, and $K_n$ is the number of questions.
However, existing checklists contain questions that ask about information unrelated to solving the original problem, are answerable from the textual input alone without inspecting the image~\cite{mathverse,emma}, or require additional calculations. To prevent such questions from confounding the assessment of visual perception, we instruct GPT-5.6 Luna~\cite{gpt56} to exclude them during checklist construction. More details are provided in supplementary \cref{sec:checklist}.

\subsection{Confidence-aware Perception Reward (CaPR)}
\label{sec:cvpr_reward}

In this section, we introduce Confidence-aware Perception Reward (CaPR), a module-specific reward for the vision encoder module. CaPR combines \textit{perception sign} and \textit{posterior confidence}. The perception sign is determined by checklist correctness, while posterior confidence calibrates the reward magnitude to attenuate signals from uncertain perceptual assessments without changing the sign.

\vspace{1ex}
\noindent\textbf{Perception sign.} We determine the perception sign from checklist correctness, independently of the main-task outcome. For problem $n$, the model generates $H$ responses to each question in the checklist $\mathcal C_n$ defined in Sec.~\ref{sec:visual_checklist}. The $h$-th responses to the $K_n$ questions collectively form a checklist trial. We compare each predicted answer in this trial with the corresponding ground-truth answer $b_{nj}$ and set $c^{\mathrm{ok}}_{nhj}=1$ if the answer is correct and $0$ otherwise. The total number of correct answers is $s_{nh}=\sum_{j=1}^{K_n}c^{\mathrm{ok}}_{nhj}$. We then compute the checklist accuracy $\hat{p}_{nh}$ and the perception sign $P_{nh}$ as:
\begin{equation}
\label{eq:perception_sign}
\hat p_{nh}=\frac{s_{nh}}{K_n},\qquad
P_{nh}=\begin{cases}
+1, & \hat p_{nh}>\tau,\\
-1, & \text{otherwise},
\end{cases}
\end{equation}
where $\tau$ is the perception threshold.

\vspace{1ex}
\noindent\textbf{Posterior confidence.}
Although the perception sign distinguishes checklist success from failure, it does not reflect how strongly the observed answers support either side of the threshold. Using empirical checklist accuracy directly as the reward magnitude may overlook differences in the number of checklist questions and weaken negative rewards as checklist accuracy decreases. We therefore introduce posterior confidence to quantify the strength of the available evidence. Let $\phi_{nh}$ denote the latent probability of correctly answering a checklist question. Assuming conditionally independent Bernoulli outcomes with a shared success probability $\phi_{nh}$ and a uniform $\mathrm{Beta}(1,1)$ prior, the posterior distribution and confidence score are formulated as follows:
\begin{equation}
\label{eq:perception_confidence}
\begin{aligned}
    & \phi_{nh}\mid s_{nh},K_n
    \sim\mathrm{Beta}\!\left(s_{nh}+1,\;K_n-s_{nh}+1\right),\\
    & p^*_{nh}=\Pr\!\left(\phi_{nh}>\tau\mid s_{nh},K_n\right),
    \quad Q_{nh}=\left|2p^*_{nh}-1\right|,
\end{aligned}
\end{equation}
where $p^*_{nh}$ denotes the posterior probability that the latent perception accuracy exceeds $\tau$, and $Q_{nh}\in[0,1]$ measures how strongly the posterior favors one side of the threshold. We combine the perception sign $P_{nh}$ and posterior confidence $Q_{nh}$ to construct a signed perception reward:
\begin{equation}
\label{eq:posterior_reward}
S^V_{nh}=P_{nh}Q_{nh}.
\end{equation}
A detailed theoretical analysis of posterior confidence is provided in supplementary \cref{sec:theoretical}.

\subsection{Module-specific advantages}
\label{sec:module_specific_advantages}

We construct two advantage signals to align optimization with each module's role. For the vision encoder module, we derive the Vision Advantage, $A^V$, from the CaPR rewards in Sec.~\ref{sec:cvpr_reward}. The projector and LLM modules share the original outcome-based GRPO advantage, $A^M$, computed from final-answer rewards.

\vspace{1ex}
\noindent\textbf{Vision advantage.}
\label{sec:grva}
We derive the vision advantage from CaPR $S^V_{nh}$ in Eq.~\eqref{eq:posterior_reward} across the $H$ checklist trials for each problem $n$. For visual reward normalization, we apply the group-relative normalization in Eq.~\eqref{eq:GRPO_advantage} using the group mean $\mu_n^V$ and sample standard deviation $\sigma_n^V$. This normalization can assign a negative advantage to a trial with a positive perception sign, or a positive advantage to a trial with a negative perception sign. To ensure that nonzero advantages agree with the perception sign, we apply a \textit{sign gate} that sets conflicting advantages to zero:
\begin{equation}
\label{eq:grva}
\widetilde{A}^V_{nh}
=\frac{S^V_{nh}-\mu_n^V}{\sigma_n^V+\epsilon},
\quad
A^{V,\mathrm{gate}}_{nh}
=
\begin{cases}
\widetilde{A}^V_{nh}, & P_{nh}\widetilde{A}^V_{nh}\geq 0,\\
0, & \text{otherwise}.
\end{cases}
\end{equation}
where $\epsilon>0$ is a small constant for numerical stability.

We further apply \textit{outcome weighting} to modulate $A^{V,\mathrm{gate}}_{nh}$ according to main-task performance. We assign larger weights to problems whose main-task rollout groups contain more balanced proportions of correct and incorrect responses, where correct answers are attainable but inconsistent. For each problem $n$, let $u_n\in[0,1]$ denote the fraction of correct final answers among its $G$ main-task responses. We use the nonnegative weight $\sqrt{u_n(1-u_n)}$, which attains its maximum of $0.5$ at $u_n=0.5$ and vanishes when all responses are correct or incorrect. The final vision advantage is computed as:
\begin{equation}
\label{eq:visual_credit_modulation}
A^V_{nh}
=A^{V,\mathrm{gate}}_{nh}\cdot\lambda_V\sqrt{u_n(1-u_n)},
\end{equation}
where $\lambda_V\geq0$ controls the overall scale of the vision advantage. This outcome weighting scales vision updates across problems without reversing the sign of $A^{V,\mathrm{gate}}_{nh}$.

\vspace{1ex}
\noindent\textbf{Outcome-based advantage.}
\label{sec:language_advantage}
Meanwhile, the projector and LLM modules retain the original outcome-based GRPO advantage. For each problem $n$, the model generates $G$ main-task responses, whose final answers are evaluated against the ground truth to obtain verifiable outcome rewards. We normalize these rewards within each problem group following Eq.~\eqref{eq:GRPO_advantage} to compute the advantage $A^M$.

\subsection{MSPO objective}
\label{sec:loss_function}

Let $\theta=(\theta_V,\theta_P,\theta_L)$ denote the trainable parameters of the vision encoder, projector, and LLM modules, respectively. We construct a main-task objective $\mathcal J_M$ and a vision objective $\mathcal J_V$ following Eq.~\eqref{eq:GRPO_objective}. These objectives are evaluated on main-task responses and separately generated checklist responses, respectively. For the vision objective, the same advantage $A^V_{nh}$ is assigned to all tokens in each of the $K_n$ checklist responses comprising trial $h$.

We maximize the following module-specific objective:
\begin{equation}
\label{eq:mspo_total_loss}
\begin{aligned}
\mathcal J_{\mathrm{MSPO}}(\theta)
={}&
\mathcal J_M\!\left(
\operatorname{sg}(\theta_V),\theta_P,\theta_L;\,A^M
\right)\\
&+
\mathcal J_V\!\left(
\theta_V,\operatorname{sg}(\theta_P),
\operatorname{sg}(\theta_L);\,A^V
\right),
\end{aligned}
\end{equation}
where $\operatorname{sg}(\cdot)$ stops gradients to the indicated parameters. The vision objective therefore updates only the vision encoder, while the main-task objective updates only the projector and LLM modules. The module-specific gradients are combined and applied in a single optimization step.

\definecolor{mainteal}{HTML}{1C716E}
\newlength{\avgcolwidth}
\newlength{\modelcolwidth}
\setlength{\modelcolwidth}{0.20\textwidth}

\newcommand{\edgepatch}[1]{%
  \noalign{%
    \hbox to \textwidth{%
      {\color{mainteal!10}%
       \rule{\dimexpr\modelcolwidth+2\tabcolsep\relax}{#1}}%
      \hfill
      {\color{mainteal!10}%
       \rule{\dimexpr\avgcolwidth+2\tabcolsep\relax}{#1}}%
    }%
  }%
}

\newcommand{\mergepatch}[1]{%
  \noalign{%
    \hbox to \textwidth{%
      {\color{mainteal!10}%
       \rule{\dimexpr\modelcolwidth+2\tabcolsep\relax}{#1}}%
      \vrule width \arrayrulewidth height #1\relax
      \hfill
      \vrule width \arrayrulewidth height #1\relax
      {\color{mainteal!10}%
       \rule{\dimexpr\avgcolwidth+2\tabcolsep\relax}{#1}}%
    }%
  }%
}

\begin{table*}[htbp]
\centering
\caption{\textbf{Main results on mathematical and general reasoning benchmarks} (Avg: mean over all seven benchmarks). We compare MSPO with existing RL-tuned baselines and report accuracy (\%). The best performance is in \textbf{bold} and the second-best is \underline{underlined}.}
\label{tab:main_results}

{\small
\setlength{\tabcolsep}{2pt}
\renewcommand{\arraystretch}{1.06}
\setlength{\extrarowheight}{1.5pt}

\begin{tabularx}{\textwidth}{
    >{\raggedright\arraybackslash\hspace*{3pt}}p{\modelcolwidth}
    |
    *{5}{>{\hsize=1.02\hsize\centering\arraybackslash}X}
    |
    *{2}{>{\hsize=1.10\hsize\centering\arraybackslash}X}
    |
    >{\hsize=0.70\hsize
      \global\avgcolwidth=\hsize
      \centering\arraybackslash}X
}

\toprule

\rowcolor{mainteal!10}
\cellcolor{mainteal!10}
& \multicolumn{5}{c|}{
    \textcolor{mainteal}{\textbf{Mathematical Reasoning}}
}
& \multicolumn{2}{c|}{
    \textcolor{mainteal}{\textbf{General Reasoning}}
}
& \cellcolor{mainteal!10} \\

\mergepatch{3pt}
\cline{2-8}
\mergepatch{2pt}

\cellcolor{mainteal!10}
\multirow{-2}{*}{Model}
& Geo3K
& WeMath
& MathVista
& MathVerse
& MathVision
& MMMU-Pro
& EMMA
& \cellcolor{mainteal!10}
  \multirow{-2}{*}{
    \textcolor{mainteal}{\textbf{Avg}}
  } \\

\edgepatch{2pt}
\specialrule{\lightrulewidth}{0pt}{\belowrulesep}

\rowcolor{gray!10}
\multicolumn{9}{c}{
    \textcolor{black}{\textit{Backbone: \textbf{Qwen2.5-VL-3B-Instruct}}}
} \\

Base model
& 27.12 & 51.84 & 55.80 & 29.44 & 18.98
& 28.38 & 23.74 & 33.61 \\

\mbox{GRPO~\cite{deepseekmath}~\textsubscript{\tiny\textcolor{black!65}{arXiv~2024}}}
& \underline{30.78} & 54.83 & 56.60 & 32.18 & \underline{20.69}
& 27.28 & 24.03 & 35.20 \\

\mbox{DAPO~\cite{dapo}~\textsubscript{\tiny\textcolor{black!65}{NeurIPS~2025}}}
& 28.62 & 53.97 & \underline{58.90} & 31.52 & 19.57
& 28.93 & \underline{25.04} & 35.22 \\

\mbox{MM-Eureka~\cite{mmeureka}~\textsubscript{\tiny\textcolor{black!65}{arXiv~2025}}}
& 27.79 & 53.74 & 57.70 & 30.30 & 18.62
& 28.50 & 23.71 & 34.34 \\

\mbox{VPPO~\cite{vppo}~\textsubscript{\tiny\textcolor{black!65}{ICLR~2026}}}
& 29.78 & \underline{55.06} & 57.40 & \underline{32.74} & 20.16
& 28.96 & 23.57 & 35.38 \\

\mbox{PAPO~\cite{papo}~\textsubscript{\tiny\textcolor{black!65}{ICLR~2026}}}
& 30.45 & 55.00 & 58.10 & 31.90 & 19.93
& \underline{29.08} & 24.03 & \underline{35.50} \\

\mbox{Perception-R1~\cite{perception_r1}~\textsubscript{\tiny\textcolor{black!65}{ICLR~2026}}}
& 29.78 & 53.56 & 57.00 & 31.32 & 19.47
& 28.61 & 23.64 & 34.77 \\

\midrule

\rowcolor{mainteal!10}
\textcolor{mainteal}{\textbf{MSPO (Ours)}}
& \textbf{31.11} & \textbf{57.30} & \textbf{60.80}
& \textbf{33.55} & \textbf{21.78}
& \textbf{30.40} & \textbf{25.50} & \textbf{37.21} \\

\midrule

\rowcolor{gray!10}
\multicolumn{9}{c}{
    \textcolor{black}{\textit{Backbone: \textbf{Qwen2.5-VL-7B-Instruct}}}
} \\

Base model
& 38.94 & 60.80 & 63.60 & 39.01 & 23.72
& 35.84 & 24.57 & 40.93 \\

\mbox{GRPO~\cite{deepseekmath}~\textsubscript{\tiny\textcolor{black!65}{arXiv~2024}}}
& \underline{41.10} & 60.06 & 67.00 & 41.12 & 23.65
& 36.42 & 24.53 & 41.98 \\

\mbox{DAPO~\cite{dapo}~\textsubscript{\tiny\textcolor{black!65}{NeurIPS~2025}}}
& 39.93 & 60.57 & \underline{67.30} & 40.25 & \underline{24.08}
& 36.27 & \underline{25.32} & 41.96 \\

\mbox{MM-Eureka~\cite{mmeureka}~\textsubscript{\tiny\textcolor{black!65}{arXiv~2025}}}
& 38.77 & 59.25 & 63.20 & 40.30 & 22.96
& 37.17 & 24.89 & 40.93 \\

\mbox{VPPO~\cite{vppo}~\textsubscript{\tiny\textcolor{black!65}{ICLR~2026}}}
& 40.77 & 60.86 & 66.50 & \underline{41.40} & 23.78
& 36.13 & 24.61 & \underline{42.01} \\

\mbox{PAPO~\cite{papo}~\textsubscript{\tiny\textcolor{black!65}{ICLR~2026}}}
& 37.60 & \textbf{64.14} & 63.50 & 38.35 & 22.93
& 35.90 & 25.00 & 41.06 \\

\mbox{Perception-R1~\cite{perception_r1}~\textsubscript{\tiny\textcolor{black!65}{ICLR~2026}}}
& 39.77 & 60.92 & 63.60 & 39.80 & 23.06
& \underline{37.20} & 24.53 & 41.27 \\

\midrule

\rowcolor{mainteal!10}
\textcolor{mainteal}{\textbf{MSPO (Ours)}}
& \textbf{42.26} & \underline{62.76} & \textbf{68.80}
& \textbf{42.97} & \textbf{26.45}
& \textbf{38.87} & \textbf{26.11} & \textbf{44.03} \\

\bottomrule
\end{tabularx}
}
\par
\end{table*}

\section{Experiments}
\subsection{Experimental setup}

\noindent\textbf{Datasets: }
We train our model on the official training split of Geometry3K (Geo3K)~\cite{geo3k}. For evaluation, we use five mathematical reasoning benchmarks, including the Geo3K test set, WeMath~\cite{wemath}, MathVista~\cite{mathvista}, MathVerse~\cite{mathverse}, and MathVision~\cite{mathvision}. We further evaluate general reasoning on the Vision split of MMMU-Pro~\cite{mmmupro} and on EMMA~\cite{emma}, which assess multidisciplinary and cross-modal reasoning, respectively. Additionally, we assess general Visual Question Answering (VQA) performance on A-OKVQA\textsubscript{\tiny MC}~\cite{aokvqa} validation and SimpleVQA-EN~\cite{simplevqa}.

\vspace{1ex}
\noindent\textbf{Implementation details: }
We use Qwen2.5-VL-3B/7B-Instruct~\cite{qwen2_5} as our backbones. We fine-tune the vision encoder, projector, and LLM using LoRA~\cite{lora} (rank 64, alpha 256) and AdamW~\cite{adamw} with a learning rate of $5\times10^{-6}$, while keeping all pretrained parameters frozen. We train for one epoch with a batch size of 128 problems. During training, we sample $G=12$ main-task responses per problem, each capped at 2,048 tokens, and $H=12$ responses per checklist question. For MSPO, we set the perception threshold $\tau=0.7$ and vision advantage scale $\lambda_V=0.9$.
More details are provided in supplementary \cref{sec:implementation}.

\vspace{1ex}
\noindent\textbf{Comparison baselines: }
We compare MSPO against six RL-tuned baselines, including GRPO~\cite{deepseekmath}, DAPO~\cite{dapo}, MM-Eureka~\cite{mmeureka}, VPPO~\cite{vppo}, PAPO~\cite{papo}, and Perception-R1~\cite{perception_r1}. We train all baselines on Geo3K using the same backbones and settings as MSPO. PEARL~\cite{pearl} and MoCA~\cite{moca} are omitted as training code is unavailable.

\subsection{Main results}
\cref{tab:main_results} presents the main experimental results on the Qwen2.5-VL-3B and Qwen2.5-VL-7B backbones.
On the 3B model, MSPO achieves the highest average accuracy of 37.21\%, improving over GRPO by 2.01 percentage points. It also outperforms the second-best model, PAPO, by 1.71 percentage points. When the backbone is scaled from 3B to 7B, MSPO maintains its lead with an average accuracy of 44.03\%. It outperforms the second-best model, VPPO, and PAPO by 2.02 and 2.97 percentage points, respectively. While baseline rankings vary across benchmarks, MSPO ranks first on all seven benchmarks at 3B and six of seven at 7B, placing second to PAPO only on WeMath at 7B. Beyond mathematical reasoning, MSPO achieves the highest accuracy on MMMU-Pro and EMMA at both scales despite being trained exclusively on Geo3K, suggesting its benefits extend to general reasoning. Overall, these comparisons support the effectiveness of guiding the vision encoder with perceptual feedback while retaining outcome-based updates for the projector and LLM.

\begin{table*}[t]
    \centering
    \caption{\textbf{Ablation study of MSPO configurations on seven mathematical and general reasoning benchmarks.} We examine objective assignment across modules, compare confidence estimates and vision advantage formulations, and ablate individual components.}
    \label{tab:ablation_all}

    {\small
    \setlength{\tabcolsep}{2.5pt}
    \renewcommand{\arraystretch}{1.12}
    \setlength{\extrarowheight}{1.5pt}

    \begin{tabularx}{\textwidth}{
        >{\hspace{3pt}}l
        *{5}{>{\hsize=0.99\hsize\centering\arraybackslash}X}
        *{2}{>{\hsize=1.15\hsize\centering\arraybackslash}X}
        >{\hsize=0.75\hsize\centering\arraybackslash}X
    }
        \toprule
        Configuration
        & Geo3K
        & WeMath
        & MathVista
        & MathVerse
        & MathVision
        & \mbox{MMMU-Pro}
        & \mbox{EMMA}
        & Avg \\
        \midrule

        \rowcolor{gray!10}
        \multicolumn{9}{c}{\textit{\textbf{Objective assignment across modules}}} \\
        Same objectives for all modules
        & 28.12 & 51.49 & 54.20 & 32.06 & 18.65 & 29.31 & 24.18 & 34.00 \\
        \midrule

        \rowcolor{gray!10}
        \multicolumn{9}{c}{\textit{\textbf{Confidence estimation}}} \\
        Checklist correct-answer rate
        & 29.12 & 56.95 & 58.90 & 32.23 & 21.38 & 30.14 & 25.11 & 36.26 \\
        \midrule

        \rowcolor{gray!10}
        \multicolumn{9}{c}{\textit{\textbf{Vision advantage formulation}}} \\
        Perception-R1-reward based $A^V$
        & 30.95 & 56.49 & 58.50 & 33.53 & 21.05 & 29.08 & 24.07 & 36.24 \\
        \midrule

        \rowcolor{gray!10}
        \multicolumn{9}{c}{\textit{\textbf{Component ablations}}} \\
        w/o Posterior Confidence
        & 29.78 & 55.69 & 58.40 & 33.35 & 21.12 & 29.34 & 24.64 & 36.05 \\
        w/o Visual reward normalization
        & 30.12 & 56.21 & 59.60 & 32.82 & 21.25 & 29.68 & 24.10 & 36.25 \\
        w/o Outcome weighting
        & 29.78 & 57.13 & 58.90 & 32.49 & 21.74 & 28.96 & 25.14 & 36.31 \\
        w/o Sign gate
        & 30.78 & 56.90 & 59.60 & 33.12 & 20.95 & 29.48 & 24.25 & 36.44 \\
        \midrule

        \rowcolor{mainteal!7}
        \textcolor{mainteal}{\textbf{MSPO (Ours)}}
        & \textbf{31.11} & \textbf{57.30} & \textbf{60.80} & \textbf{33.55} & \textbf{21.78} & \textbf{30.40} & \textbf{25.50} & \textbf{37.21} \\
        \bottomrule
    \end{tabularx}
    }
\end{table*}

{\small
\begin{table}[t]
    \centering
    \caption{\textbf{Ablation study of assigning $A^V$ or $A^M$ to the projector module.}}
    \label{tab:module_grouping}

    {\small
    \setlength{\tabcolsep}{3pt}
    \renewcommand{\arraystretch}{1.12}
    \setlength{\extrarowheight}{0pt}
    \setlength{\aboverulesep}{0.5ex}
    \setlength{\belowrulesep}{0.5ex}
    \newsavebox{\modulegroupingtablebox}

    \begin{lrbox}{\modulegroupingtablebox}
    \begin{tabularx}{\linewidth}{
        *{4}{>{\centering\arraybackslash}X}
    }
        \toprule
        Vision
        & \multirow[c]{2}{*}{Projector}
        & \multirow[c]{2}{*}{LLM}
        & \multirow[c]{2}{*}{Avg} \\
        encoder & & & \\
        \midrule

        $A^V$ & $A^V$ & $A^M$
        & 35.36 \\
        \midrule

        \rowcolor{mainteal!7}
        {\mbox{\textbf{$A^V$}}}
        & \textcolor{mainteal}{\mbox{\textbf{$A^M$}}}
        & {\mbox{\textbf{$A^M$}}}
        & \textbf{37.21} \\
        \bottomrule
    \end{tabularx}
    \end{lrbox}

    \usebox{\modulegroupingtablebox}%
    \makebox[0pt][l]{%
        \hspace*{-0.25\linewidth}%
        \raisebox{-\dp\modulegroupingtablebox}{%
            \rule{\arrayrulewidth}{%
                \dimexpr\ht\modulegroupingtablebox
                +\dp\modulegroupingtablebox\relax
            }%
        }%
    }%
    }
\end{table}
}

\subsection{Ablation study}
All ablation studies of MSPO are conducted using the Qwen2.5-VL-3B backbone, with accuracy averaged across seven mathematical and general reasoning benchmarks.

\vspace{1ex}
\noindent\textbf{Ablation of MSPO Configurations:}
\cref{tab:ablation_all} presents ablation results for MSPO configurations. Applying both the checklist and main-task objectives to all modules in a single optimizer step per batch lowers average accuracy. This supports assigning the checklist objective to the vision encoder and the main-task objective to the projector and LLM. Replacing posterior confidence with the checklist correct-answer rate while retaining perception signs reduces performance. Constructing $A^V$ from Perception-R1~\cite{perception_r1} rewards instead of CaPR degrades performance, supporting CaPR's formulation. Details of this comparison are provided in supplementary \cref{sec:perception_r1_advantage}. Removing posterior confidence retains only the perception sign as the reward, lowering average accuracy and highlighting the benefit of confidence-based reward scaling. Removing visual reward normalization uses each checklist trial's CaPR without comparison to other trials, reducing average accuracy and highlighting the benefit of group-relative comparison. Removing outcome weighting degrades performance, supporting modulation of vision updates by main-task performance. Removing the sign gate lowers average accuracy, supporting alignment of advantage signs with perceptual correctness.

\vspace{1ex}
\noindent\textbf{Advantage assignment to the projector module:}
To determine which advantage better guides projector optimization, we compare $A^V$ and $A^M$ with other module assignments fixed. As shown in \cref{tab:module_grouping}, assigning $A^M$ rather than $A^V$ improves overall average accuracy from 35.36\% to 37.21\%. This supports sharing the outcome-based advantage between the projector and LLM, consistent with their shared response-generation objective in LLaVA's visual instruction tuning~\cite{llava}.

{\small
\begin{table}[tbp]
    \centering
    \caption{\textbf{Sensitivity study of MSPO's perception threshold $\tau$ and vision advantage scale $\lambda_V$.}}
    \label{tab:tau_ablation}

    {\small
    \setlength{\tabcolsep}{4pt}
    \renewcommand{\arraystretch}{1.12}
    \newsavebox{\hyperparamtablebox}

    \begin{lrbox}{\hyperparamtablebox}
    \begin{tabularx}{\linewidth}{
        *{3}{>{\centering\arraybackslash}X}
    }
        \toprule
        $\tau$
        & $\lambda_V$
        & Avg \\
        \midrule
        0.9 & 1.0 & 36.71 \\
        0.9 & 0.9 & 36.67 \\
        0.8 & 1.0 & 36.83 \\
        0.8 & 0.9 & 36.76 \\
        0.7 & 1.0 & 37.12 \\
        \midrule
        \rowcolor{mainteal!10}
        \textcolor{mainteal}{\mbox{\textbf{0.7}}} & \textcolor{mainteal}{\mbox{\textbf{0.9}}}
        & \textbf{37.21} \\
        \bottomrule
    \end{tabularx}
    \end{lrbox}

    \usebox{\hyperparamtablebox}%
    \makebox[0pt][l]{%
        \hspace*{\dimexpr-\linewidth/3\relax}%
        \raisebox{-\dp\hyperparamtablebox}{%
            \rule{\arrayrulewidth}{%
                \dimexpr\ht\hyperparamtablebox+\dp\hyperparamtablebox\relax
            }%
        }%
    }%
    }
\end{table}
}

\vspace{1ex}
\noindent\textbf{Hyperparameter sensitivity:}
\cref{tab:tau_ablation} presents MSPO's sensitivity to the perception threshold $\tau$ and vision advantage scale $\lambda_V$. Among these configurations, $\tau=0.7$ and $\lambda_V=0.9$ yield the highest average accuracy. Increasing $\tau$ from $0.7$ to $0.8$ or $0.9$ lowers accuracy, possibly because stricter thresholds can assign negative perception signs even to trials with mostly correct checklist responses. Further sensitivity analysis is provided in supplementary \cref{sec:tau_sensitivity}.

\subsection{Compatibility \& generalization}
We examine MSPO's compatibility with different policy optimization methods and evaluate its generalization to out-of-domain visual question answering tasks. All experiments in this section use the Qwen2.5-VL-3B backbone.

{\small
\begin{table}[t]
    \centering
    \caption{\textbf{Compatibility with policy optimization methods.}}
    \label{tab:compatibility}

    {\small
    \setlength{\tabcolsep}{2.5pt}
    \renewcommand{\arraystretch}{1.12}

    \begin{tabularx}{\linewidth}{
        >{\hspace{3pt}}l
        >{\centering\arraybackslash}X
        >{\columncolor{mainteal!10}\centering\arraybackslash}X
    }
        \toprule
        \multirow[c]{2}{*}{Method} & \multicolumn{2}{c}{Avg} \\
        \cmidrule(lr){2-3}
        & Native & \textcolor{mainteal}{\mbox{\textbf{+ Ours ($A^V$)}}} \\
        \midrule

        DAPO & 35.22 & \textbf{36.92} \\
        VPPO & 35.38 & \textbf{35.93} \\
        PAPO & 35.50 & \textbf{35.95} \\
        \bottomrule
    \end{tabularx}
    }
\end{table}
}

\vspace{1ex}
\noindent\textbf{Compatibility with policy optimization methods: }
\cref{tab:compatibility} evaluates the compatibility of MSPO with DAPO, VPPO, and PAPO by comparing their original performance with the results obtained using MSPO. We integrate MSPO with each method in a plug-and-play manner by updating the vision encoder with $A^V$ while preserving each method's original optimization objectives for the projector and LLM. MSPO improves average accuracy across the seven benchmarks for all three methods. These results demonstrate that module-specific credit assignment can complement existing policy optimization strategies.

{\small
\begin{table}[t]
    \centering
    \caption{\textbf{Generalization to out-of-domain VQA tasks.}}
    \label{tab:vqa}

    \setlength{\tabcolsep}{2pt}
    \renewcommand{\arraystretch}{1.12}

    \begin{tabularx}{\linewidth}{
        l
        >{\hsize=1.10\hsize\centering\arraybackslash}X
        >{\hsize=1.25\hsize\centering\arraybackslash}X
        >{\hsize=0.65\hsize\centering\arraybackslash}X
    }
        \toprule
        Model
        & \mbox{A-OKVQA\textsubscript{\tiny MC}}
        & \mbox{SimpleVQA-EN}
        & Avg \\
        \midrule
        \hspace{3pt}GRPO & 83.84 & 37.91 & 60.88 \\
        \hspace{3pt}DAPO & 84.02 & 38.10 & 61.06 \\
        \rowcolor{mainteal!10}
        \hspace{3pt}\textcolor{mainteal}{\mbox{\textbf{MSPO (Ours)}}}
        & \textbf{84.37} & \textbf{38.40} & \textbf{61.39} \\
        \bottomrule
    \end{tabularx}
\end{table}
}

\vspace{1ex}
\noindent{\textbf{Generalization to VQA (out-of-domain): }}
\label{section:VQA}
To assess MSPO's generalization beyond mathematical and general reasoning benchmarks, we evaluate its performance on out-of-domain VQA tasks. As shown in \cref{tab:vqa}, MSPO achieves 84.37\% accuracy on A-OKVQA\textsubscript{\tiny MC} and 38.40\% on SimpleVQA-EN, outperforming GRPO and DAPO. These results demonstrate that MSPO is also effective on general VQA tasks despite being trained exclusively on Geo3K. This generalization is consistent with MSPO's module-specific design, which uses perceptual feedback to guide vision learning toward representations useful across tasks.

{\small
\begin{figure}[t]
    \centering
    \includegraphics[width=0.9\linewidth]{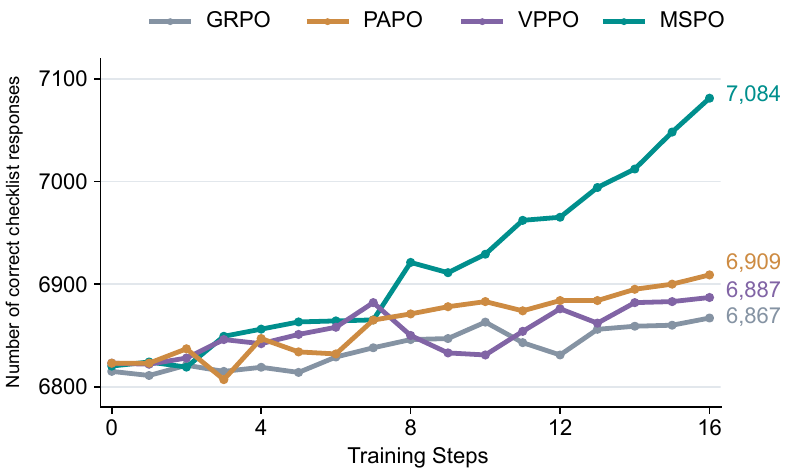}
    \caption{\textbf{Perception correctness during training.} Curves show the total number of correct checklist responses on the Geo3K validation set, with 12 responses sampled per question.}
    \label{fig:perception_progress}
\end{figure}
}

{\small
\begin{figure*}[t]
    \centering
    \includegraphics[width=\textwidth]{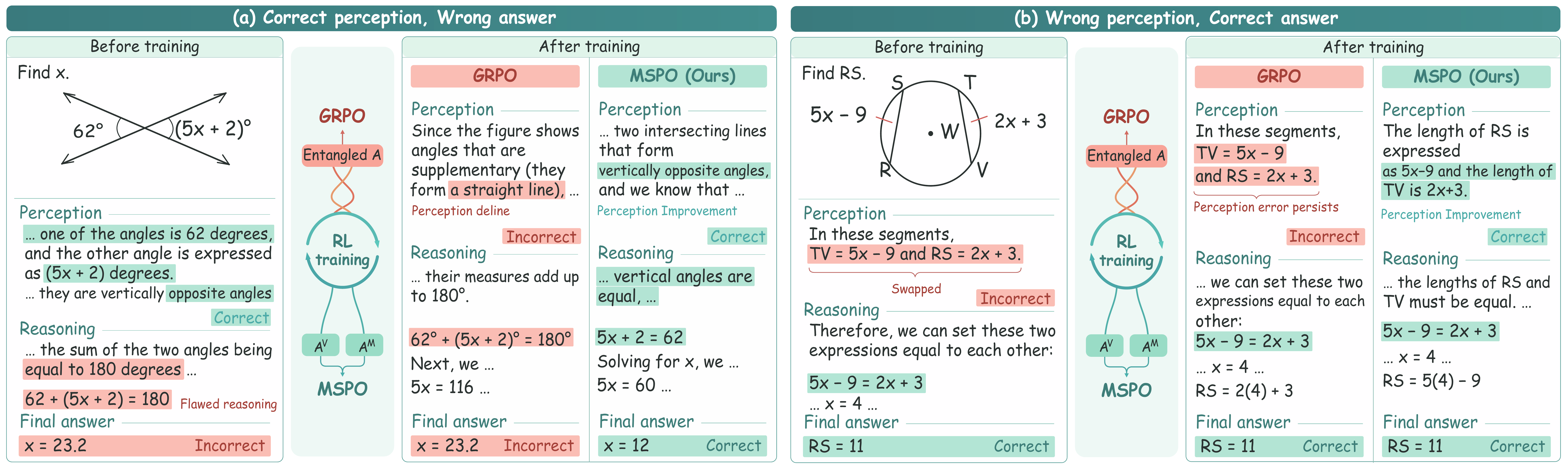}
    \caption{\textbf{Qualitative comparison of GRPO and MSPO in addressing Module-Level Credit Entanglement.} Responses before and after training for two initial conditions: (a) correct perception with an incorrect answer and (b) incorrect perception with a correct answer.}
    \label{fig:training_outcomes}
\end{figure*}
}

\subsection{Perception improvement during training}

To assess whether MSPO improves visual perception alongside main-task performance, we evaluate Qwen2.5-VL-3B-Instruct models trained with GRPO, PAPO, VPPO, and MSPO at successive checkpoints using the same perception checklists on the Geo3K validation set. At each checkpoint, we sample 12 responses per checklist question and count correct responses independently of final-answer correctness. As shown in \cref{fig:perception_progress}, MSPO yields more correct responses than the baselines from step 8 onward, reaching 7,084 correct checklist responses at the final step, 175 more than PAPO. These results suggest improved task-relevant perception on held-out problems.

{\small
\begin{figure}[t]
    \centering
    \includegraphics[height=92bp]{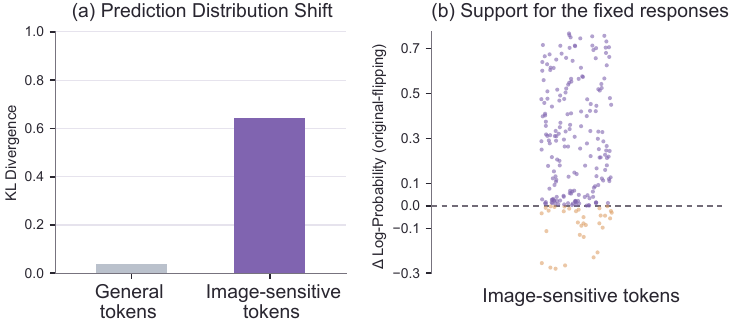}
    \captionsetup{hypcap=false}
    \caption{\textbf{Changes in token predictions under checklist flipping.} Image-sensitive tokens are the top 3\% of response tokens ranked by image sensitivity. (a): Mean KL divergence between next-token prediction distributions under original and flipped checklists by token group. (b): Per-problem mean log-probability differences for image-sensitive tokens in fixed responses, computed as original minus flipped.}
    \label{fig:stability_placeholder}
\end{figure}
}

\subsection{Perception-reasoning alignment analysis}
\cref{fig:stability_placeholder} assesses the consistency between the perceptual information in the model's checklist answers and its original main-task responses. With Geo3K validation responses fixed and images removed, we compare token prediction distributions and log-probabilities under original checklist answers and flipped answers containing conflicting visual information. We focus on image-sensitive tokens (detailed in supplementary \cref{sec:image_sensitive_token_selection}). In (a), checklist flipping induces over 17 times greater mean KL divergence at image-sensitive tokens than at remaining tokens, indicating greater sensitivity to changes in checklist-provided visual information. In (b), the original checklist assigns approximately 46\% higher probability to these tokens in terms of the geometric mean and yields higher mean token log-probabilities in 239 of 282 problems (84.8\%).

\subsection{Perception-advantage alignment analysis}
\cref{fig:perception_advantage_alignment} shows closer correspondence between perceptual assessments and vision-encoder advantage signs under MSPO than under GRPO. Qwen3.8-27B~\cite{qwen3_8} judges whether main-task training responses correctly identify task-relevant visual information. For each problem-step group, we compare judged perception accuracy across 12 main-task rollouts with the positive advantage rate across 12 values, using $\hat{A}$ for GRPO in Eq.~\eqref{eq:GRPO_advantage} and $A^V$ for MSPO in Eq.~\eqref{eq:visual_credit_modulation} (detailed in supplementary \cref{sec:perception_advantage_alignment_details}). MSPO achieves a Pearson correlation of 0.72 versus 0.43 for GRPO and reduces the mean absolute difference between these proportions from 31.62 to 26.73 percentage points. These results suggest that MSPO's vision advantages more closely reflect assessed perceptual correctness. 

{\small
\begin{figure}[t]
    \centering

    \includegraphics[width=\linewidth,height=92bp,keepaspectratio]{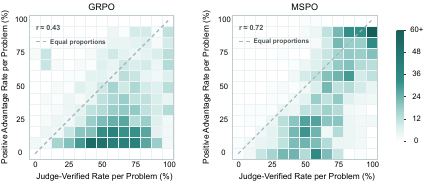}
    \captionsetup{hypcap=false}
    \caption{\textbf{Perception-Advantage Alignment.} Problem-level correspondence between judged perception accuracy and positive advantage rates over 12 samples, excluding all-zero groups. Colors show group counts; the diagonal denotes equal proportions.}
    \label{fig:perception_advantage_alignment}
\end{figure}
}

\subsection{Qualitative analysis}
\cref{fig:training_outcomes} compares GRPO and MSPO in addressing Module-Level Credit Entanglement. In case (a), before training, correct perception is followed by flawed reasoning and an incorrect answer. GRPO's shared negative advantage also penalizes the vision encoder, and perception deteriorates after training. MSPO preserves perception and corrects the answer. In case (b), before training, faulty perception nevertheless yields a correct answer. This error persists after GRPO training but is corrected by MSPO. These cases demonstrate the benefit of module-specific policy optimization in aligning learning signals with each module's role, preserving correct perception and correcting perceptual errors to support improved final-answer accuracy.

\section{Conclusion}
In this work, we introduce Module-Specific Policy Optimization (MSPO), the first framework to address Module-Level Credit Entanglement through module-specific advantages in multimodal reinforcement learning. MSPO computes the Confidence-aware Perception Reward (CaPR) from checklist responses, with a correctness-based Perception Sign determining reward direction and posterior confidence scaling reward magnitude. The resulting vision advantages guide the vision encoder, while the projector and LLM retain outcome-based GRPO updates. Experiments on seven mathematical and general reasoning benchmarks demonstrate that MSPO outperforms existing RL-tuned models on average at both the 3B and 7B scales.

\clearpage
\section*{LLM Usage Statement}
OpenAI Codex was used to assist with code implementation and verification. ChatGPT (OpenAI) was used for LaTeX table formatting and manuscript language refinement, including improvements to grammar, wording, and clarity. The OpenAI GPT API was used to construct the checklist dataset. These tools were not used to develop the core research ideas or draft the main content of the manuscript. The authors were solely responsible for the substantive manuscript content and reviewed and verified all AI-assisted outputs.

{
    \small
    \bibliographystyle{ieeenat_fullname}
    \bibliography{main}
}
\clearpage
\maketitlesupplementary
\appendix
\providecolor{mainteal}{HTML}{1C716E}

\noindent{This supplementary material is organized as follows:}
\vspace{1ex}
\begin{itemize}
  \item \textbf{Section A (Algorithm Implementation).} Pseudocode for rollout collection, advantage computation, and module-specific policy updates.
  \item \textbf{Section B (Additional Information).} Datasets, implementation settings, checklist construction, and analysis protocols.
  \item \textbf{Section C (Additional Experimental Analysis).} Sensitivity, ablations, baseline comparisons, training cost, and robustness.
  \item \textbf{Section D (Theoretical Analysis).} Analysis of the reward and advantage formulation.
  \item \textbf{Section E (Additional Qualitative Results).} Examples of perception and reasoning behavior.
  \item \textbf{Section F (Limitations and Future Work).} Limitations of MSPO and directions for future research.
\end{itemize}

\section{Algorithm implementation}
\label{sec:Algorithm}

\cref{algorithm:mspo} presents one MSPO training update in three phases: rollout collection, module-specific advantage computation, and module-specific optimization. Gradients from the checklist and main-task objectives are applied to their designated modules in a single optimizer step.

\begin{algorithm}[htbp]
\caption{Module-Specific Policy Optimization}
\label{algorithm:mspo}
\footnotesize

\textbf{Input:} Trainable parameters
$\theta=(\theta_V,\theta_P,\theta_L)$ and training set $\mathcal D$. \\[1pt]
\textbf{Parameters:} $B$ retained groups, $G$ main-task responses,
$H$ checklist trials, perception threshold $\tau$,
and vision scale $\lambda_V$. \\[1pt]
\textbf{Output:} Updated parameters $\theta$.

\begin{algorithmic}[1]

\Statex $\triangleright$\ \textbf{Phase 1:} Rollout collection

\State $\theta_{\mathrm{old}}\gets\theta,\quad
        \mathcal B\gets\varnothing,\quad
        \mathcal B_V\gets\varnothing$

\While{$|\mathcal B|<B$}
    \State $(I_n,q_n,y_n,\mathcal C_n)\sim\mathcal D$
    \State $o_{ng}\sim
        \pi_{\theta_{\mathrm{old}}}(\cdot\mid I_n,q_n)$
        for $g=1,\ldots,G$
    \State $R^M_{ng}\gets
        \operatorname{Grade}(o_{ng},y_n)$ for each $g$
    \Statex \hspace{\algorithmicindent}
        $+1$ for a correct final answer, $-1$ otherwise
    \State $u_n\gets
        G^{-1}\sum_{g=1}^{G}\mathbf 1[R^M_{ng}>0]$
    \If{$0<u_n<1$}
        \State Retain group $n$ in $\mathcal B$
    \EndIf
\EndWhile

\Statex $\triangleright$\ \textbf{Phase 2:} Module-specific advantages

\Statex $\mathsf Z_a(x_a)=
    \dfrac{x_a-\operatorname{mean}_a x_a}
    {\operatorname{std}_a x_a+\epsilon}$,
    $\epsilon=10^{-6}$
\Statex Constant rewards yield zero normalized advantages.

\For{each $n\in\mathcal B$}

    \Statex \hspace{\algorithmicindent}
        \textit{Outcome-based advantage}
    \State $A^M_{ng}\gets\mathsf Z_g(R^M_{ng})$

    \If{$\mathcal C_n$ is missing, empty, or invalid}
        \State \textbf{continue}
    \EndIf

    \Statex \hspace{\algorithmicindent}
        $\mathcal C_n=\{(c_{nj},b_{nj})\}_{j=1}^{K_n}$,
        with $K_n$ checklist items

    \State $\mathcal B_V\gets\mathcal B_V\cup\{n\}$

    \State $o^V_{nhj}\sim
        \pi_{\theta_{\mathrm{old}}}(\cdot\mid I_n,c_{nj})$
    \Statex \hspace{\algorithmicindent}
        for $h=1,\ldots,H$ and $j=1,\ldots,K_n$
    \Statex \hspace{\algorithmicindent}
        Trial $h$ collects the $h$-th response to each item.

    \Statex \hspace{\algorithmicindent}
        \textit{Perception sign}
    \State $c^{\mathrm{ok}}_{nhj}\gets
        \mathbf 1[\operatorname{Match}(o^V_{nhj},b_{nj})]$
    \State $s_{nh}\gets
        \sum_{j=1}^{K_n}c^{\mathrm{ok}}_{nhj}$
    \State $P_{nh}\gets
        2\mathbf 1[s_{nh}/K_n>\tau]-1$

    \Statex \hspace{\algorithmicindent}
        \textit{Posterior confidence and CaPR}
    \State $\phi_{nh}\mid s_{nh},K_n
        \sim\mathrm{Beta}(s_{nh}+1,K_n-s_{nh}+1)$
    \Statex \hspace{\algorithmicindent}
        under a uniform $\mathrm{Beta}(1,1)$ prior
    \State $S^V_{nh}\gets
        P_{nh}\left|
        2\Pr(\phi_{nh}>\tau\mid s_{nh},K_n)-1
        \right|$

    \Statex \hspace{\algorithmicindent}
        \textit{Vision advantage}
    \State $z_{nh}\gets\mathsf Z_h(S^V_{nh})$
        across the $H$ trials
    \State $A^V_{nh}\gets
        \lambda_V\sqrt{u_n(1-u_n)}\,
        z_{nh}\,\mathbf 1[P_{nh}z_{nh}\ge0]$
    \State $A^{V,\mathrm{token}}_{nhjt}\gets A^V_{nh}$
        for all response tokens $t$

\EndFor

\Statex $\triangleright$\ \textbf{Phase 3:} Module-specific optimization

\Statex Both objectives use the GRPO formulation in Eq.~\eqref{eq:GRPO_objective}.

\State Construct $\mathcal J_M$ from main-task rollouts
    in $\mathcal B$ with $A^M$

\State $\mathcal J_V\gets0$
\If{$\mathcal B_V\ne\varnothing$}
    \State Construct $\mathcal J_V$ from checklist rollouts
        in $\mathcal B_V$ with $A^V$
\EndIf

\State $g_V\gets
    -\nabla_{\theta_V}\mathcal J_V(\theta;A^V)$
\State $(g_P,g_L)\gets
    -\nabla_{(\theta_P,\theta_L)}\mathcal J_M(\theta;A^M)$

\State $\theta\gets\operatorname{AdamW}\!\left(
    \theta,\operatorname{ClipGrad}(g_V,g_P,g_L)
    \right)$
\Statex $\operatorname{ClipGrad}$ denotes gradient clipping.

\State \Return $\theta$

\end{algorithmic}
\end{algorithm}

\section{Additional information}
\label{sec:additional_information}

\subsection{Datasets}
\cref{tab:supp_dataset_counts} summarizes the datasets and sample counts used for training and evaluation.

\vspace{1ex}
\noindent\textbf{Mathematical Reasoning:}
We use five datasets for mathematical reasoning evaluation.
\begin{itemize}
  \vspace{1ex}
  \item \textbf{Geometry3K (Geo3K)}~\cite{geo3k} contains geometry problems paired with diagrams, with official splits of 2,101 training, 300 validation, and 601 test problems. We train MSPO on the full training split. Our training perception checklists cover 1,949 problems and contain 6,628 question--answer pairs. The other 152 training problems can contribute only to the main-task objective. Our validation perception checklists contain 943 question--answer pairs and are used for perception analysis. We use the 601-problem test split for final-answer evaluation. For Perception-R1~\cite{perception_r1}, we use the subset of 1,442 training problems with publicly available perception annotations. All other RL-tuned baselines (GRPO, DAPO, MM-Eureka, VPPO, and PAPO) use the full training split of 2,101 problems. To address diagram--text inconsistencies, ambiguous conditions, or incorrect answers, we revise question statements or reference answers in 32 of the 2,101 training problems. MSPO and all baselines use this revised version of Geo3K.

  \vspace{1ex}
  \item \textbf{WeMath}~\cite{wemath} contains visual mathematics problems spanning 67 hierarchically organized knowledge concepts. It pairs composite problems with subproblems targeting their constituent concepts, allowing assessment of both individual concepts and their combination. We use the 1,740-example \texttt{testmini} split for evaluation.

  \vspace{1ex}
  \item \textbf{MathVista}~\cite{mathvista} contains mathematical questions involving diverse visual inputs, including diagrams, plots, and tables. Its tasks require geometric reasoning and quantitative interpretation of visual information. We use the 1,000-example \texttt{testmini} split for evaluation.

  \vspace{1ex}
  \item \textbf{MathVerse}~\cite{mathverse} provides six versions of each visual mathematics problem by varying the information available in the text and image. We use the 3,940-example \texttt{testmini} split for evaluation, comprising five image-containing versions of each of 788 source problems and excluding the text-only version.

  \vspace{1ex}
  \item \textbf{MathVision}~\cite{mathvision} contains visually grounded problems from mathematics competitions, spanning 16 mathematical disciplines and five difficulty levels. We use the full test set of 3,040 problems for evaluation.
\end{itemize}

\vspace{1ex}
\noindent\textbf{General Reasoning:}
We use two datasets to assess generalization beyond the geometry training domain.
\begin{itemize}
  \vspace{1ex}
  \item \textbf{MMMU-Pro}~\cite{mmmupro} contains college-level multimodal questions across disciplines. It filters out questions answerable without visual input and expands answer choices to reduce guessing. We use the 1,730-question test set under the Vision setting, where the question and answer options are embedded in the image.

  \vspace{1ex}
  \item \textbf{EMMA}~\cite{emma} contains problems in mathematics, physics, chemistry, and coding that require joint reasoning over visual and textual information. We use all 2,788 problems for evaluation.
\end{itemize}

\vspace{1ex}
\noindent\textbf{Visual Question Answering (VQA):}
We use two datasets to assess generalization to knowledge-based visual question answering.
\begin{itemize}
  \vspace{1ex}
  \item \textbf{A-OKVQA\textsubscript{\tiny MC}}~\cite{aokvqa} contains questions requiring visual understanding together with commonsense and world knowledge. We use the 1,145-question validation split for multiple-choice evaluation.

  \vspace{1ex}
  \item \textbf{SimpleVQA-EN}~\cite{simplevqa} is the English subset of SimpleVQA, which assesses factual accuracy through visual questions with short, open-ended answers. We use all 1,013 English questions for evaluation.
\end{itemize}

\begin{table}[!t]
    \centering
    \caption{\textbf{Datasets and sample counts.}
    Checklist counts refer to question--answer pairs; other counts refer to problems.}
    \label{tab:supp_dataset_counts}

    \small
    \setlength{\tabcolsep}{3pt}
    \renewcommand{\arraystretch}{1.12}

    \begin{tabularx}{\linewidth}{
        @{}l
        >{\raggedright\arraybackslash}X
        >{\centering\arraybackslash}p{0.14\linewidth}@{}
    }
        \toprule
        Dataset & Split or use & Count \\
        \midrule

        \rowcolor{gray!10}
        \multicolumn{3}{c}{\textit{Training}} \\
        Geometry3K & Main-task training (MSPO) & 2,101 \\
        Geometry3K & Vision training (MSPO) & 1,949 \\
        Geometry3K & Perception-R1 training & 1,442 \\
        \midrule

        \rowcolor{gray!10}
        \multicolumn{3}{c}{\textit{Perception checklists}} \\
        Geometry3K & Training & 6,628 \\
        Geometry3K & Validation & 943 \\
        \midrule

        \rowcolor{gray!10}
        \multicolumn{3}{c}{\textit{Mathematical reasoning}} \\
        Geometry3K & Test & 601 \\
        WeMath & \texttt{testmini} & 1,740 \\
        MathVista & \texttt{testmini} & 1,000 \\
        MathVerse & \texttt{testmini} & 3,940 \\
        MathVision & Test & 3,040 \\
        \midrule

        \rowcolor{gray!10}
        \multicolumn{3}{c}{\textit{General reasoning}} \\
        MMMU-Pro & Test (Vision setting) & 1,730 \\
        EMMA & Full benchmark & 2,788 \\
        \midrule

        \rowcolor{gray!10}
        \multicolumn{3}{c}{\textit{VQA}} \\
        A-OKVQA\textsubscript{\tiny MC} & Validation (multiple choice) & 1,145 \\
        SimpleVQA-EN & English evaluation set & 1,013 \\
        \bottomrule
    \end{tabularx}
\end{table}

\subsection{Implementation details}
\label{sec:implementation}

We use Qwen2.5-VL-3B/7B-Instruct~\cite{qwen2_5} as our backbones and generate training rollouts and evaluation responses with vLLM. We train each model for one epoch (16 optimizer steps) with 128 problem groups per step, using BF16 and gradient checkpointing. We apply LoRA~\cite{lora} ($r=64$, $\alpha=256$) to all linear layers in the vision encoder, projector, and LLM except the final output head, while keeping all pretrained parameters frozen.
During training, we sample $G=12$ main-task responses per problem and $H=12$ responses per checklist item. Responses with the same sample index across checklist items form one checklist trial. A rule-based verifier assigns a main-task reward of $+1$ for a correct boxed answer and $-1$ otherwise. For MSPO, we use DAPO's dynamic sampling~\cite{dapo} to retain main-task rollout groups containing both correct and incorrect responses. We set (temperature, top-$p$) to $(1.0, 1.0)$ for training and $(0, 0.95)$ for math evaluation. We limit main-task responses to 2,048 tokens and checklist responses to 64 tokens to reduce checklist generation overhead.
We set the perception threshold to $\tau=0.7$ and the vision advantage scale to $\lambda_V=0.9$. The main-task objective $\mathcal{J}_M$ and the vision objective $\mathcal{J}_V$ both use a PPO clipping range of $[0.8,1.2]$ and a KL penalty coefficient of $0.01$. For the main task, we average the loss over generated response tokens within each microbatch, then average the resulting microbatch losses. For the vision objective, we first average the loss over tokens in each checklist response. We then average over checklist items within each trial, over trials within each problem, and finally over the problems contributing to the vision objective. We use AdamW~\cite{adamw} with a constant learning rate of $5\times10^{-6}$, $(\beta_1,\beta_2)=(0.9,0.999)$, and a weight decay of $0.01$. We clip the global gradient norm to $1.0$. All experiments are conducted on a system with a 24-core AMD EPYC 7413 CPU (2.65~GHz base clock), using PyTorch~2.8.0 and either one NVIDIA RTX A6000 GPU for the 3B model or two for the 7B model.

\subsection{Checklist construction}
\label{sec:checklist}
We construct perception checklists to assess whether a model recognizes the visual evidence needed to solve a geometry problem. PEARL's checklist-generation prompt~\cite{pearl} allows items derived from facts stated in the problem or obtained through one-step inference. For MSPO, we identify a valid solution path and retain only items that test visual facts directly observable in the diagram and needed for that solution. We exclude items that can be answered from the problem text alone or that require additional reasoning.

\begin{figure*}[tp]
    \centering
    \includegraphics[
        width=\linewidth,
        height=0.90\textheight,
        keepaspectratio
    ]{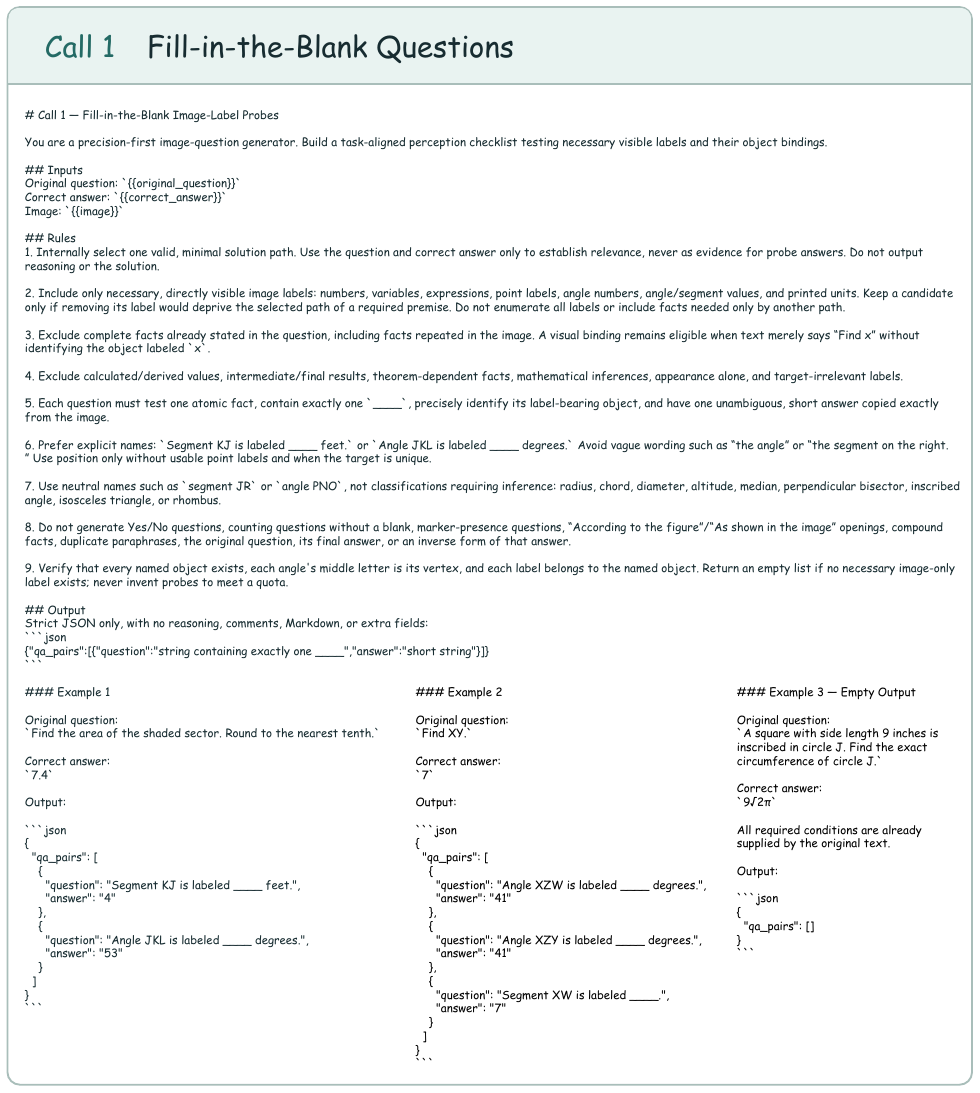}
    \caption{\textbf{Perception checklist generation prompt (Call~1).}}
    \label{fig:supp_checklist_examples}
\end{figure*}

\begin{figure*}[tp]
    \centering
    \includegraphics[
        width=\linewidth,
        height=0.90\textheight,
        keepaspectratio
    ]{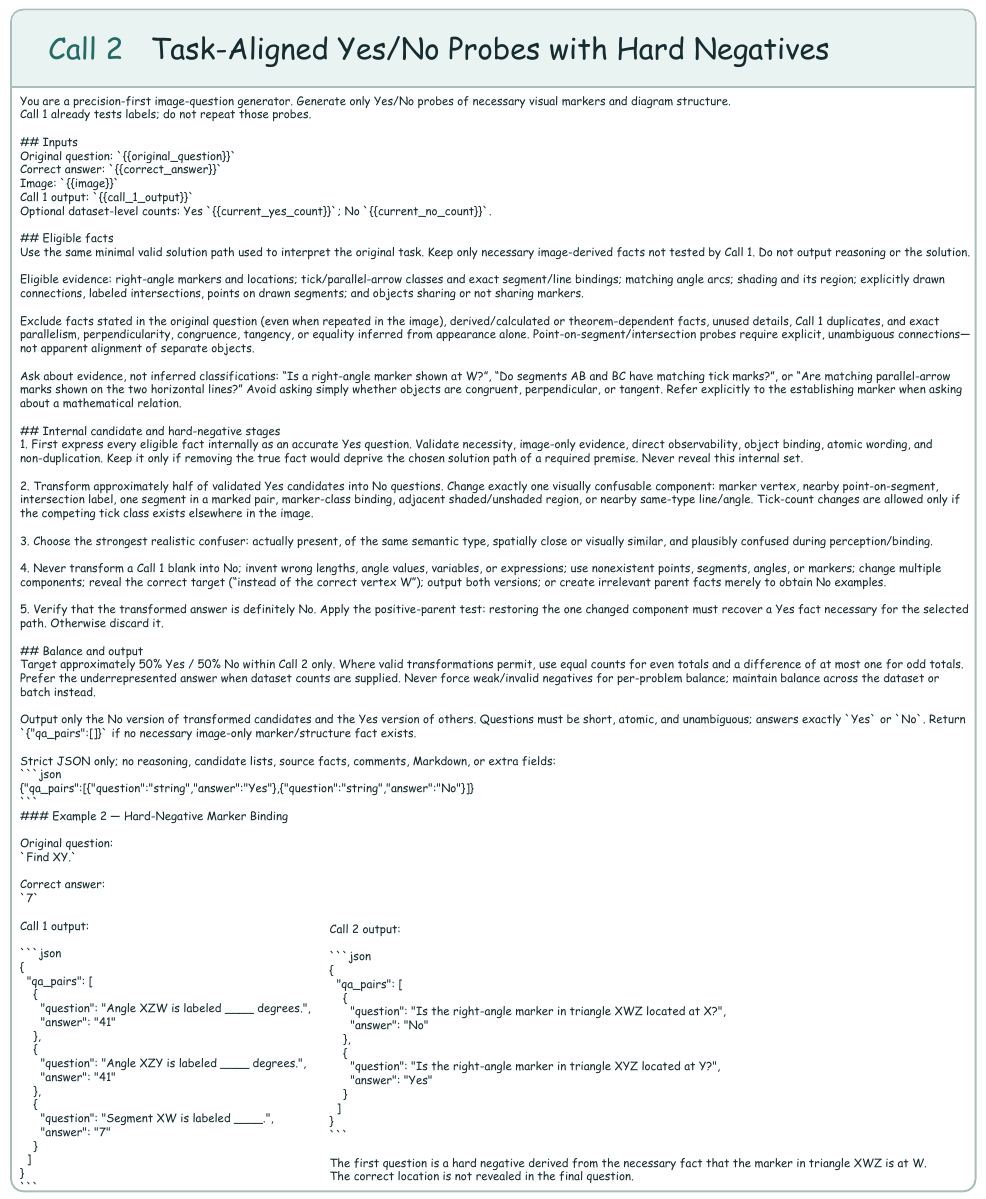}
    \caption{\textbf{Perception checklist generation prompt (Call~2).}}
    \label{fig:supp_checklist_call2}
\end{figure*}

Given the original geometry question, its diagram, and its correct answer, we use GPT-5.6 Luna~\cite{gpt56} to generate checklist question--answer pairs in two calls. Call~1 (\cref{fig:supp_checklist_examples}) generates fill-in-the-blank questions about labels needed for the solution and the objects they refer to. Call~2 (\cref{fig:supp_checklist_call2}) generates Yes/No questions about visual markings and spatial relations. To limit bias toward either response, we aim for roughly equal numbers of questions whose correct answer is Yes and questions whose correct answer is No. We allow empty checklists when no item meets the selection criteria.

\begin{figure*}[t]
    \centering
    \includegraphics[width=\linewidth]{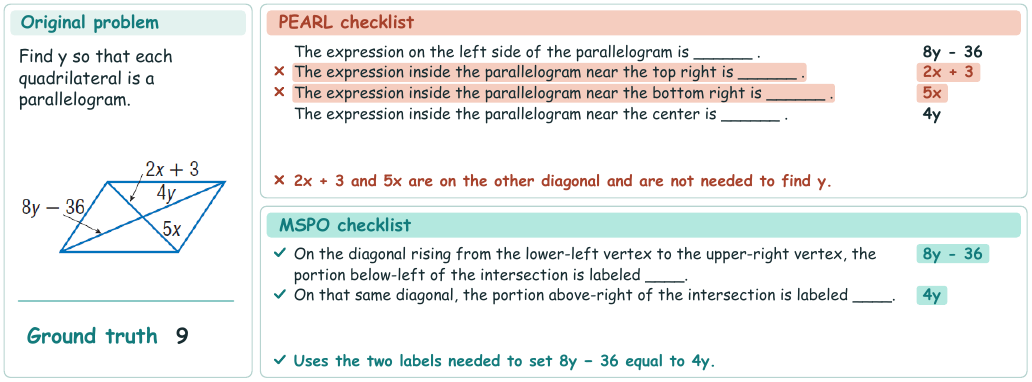}
    \caption{\textbf{Task relevance.}
    The task-aligned checklist retains the two labels on the diagonal needed to determine $y$.}
    \label{fig:supp_checklist_comparison}
\end{figure*}

\begin{figure*}[t]
    \centering
    \includegraphics[width=\linewidth]{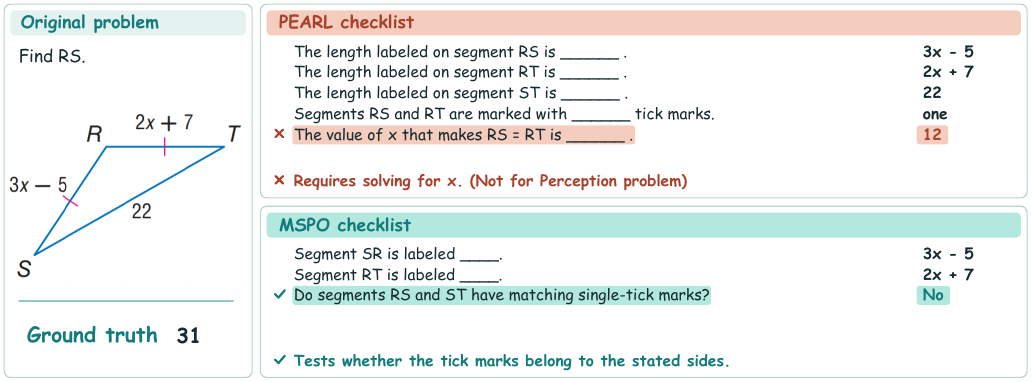}
    \caption{\textbf{Perception versus computation.}
    The task-aligned checklist checks the placement of tick marks on the sides of the triangle.}
    \label{fig:supp_checklist_comparison_2}
\end{figure*}

\begin{figure*}[t]
    \centering
    \includegraphics[width=\linewidth]{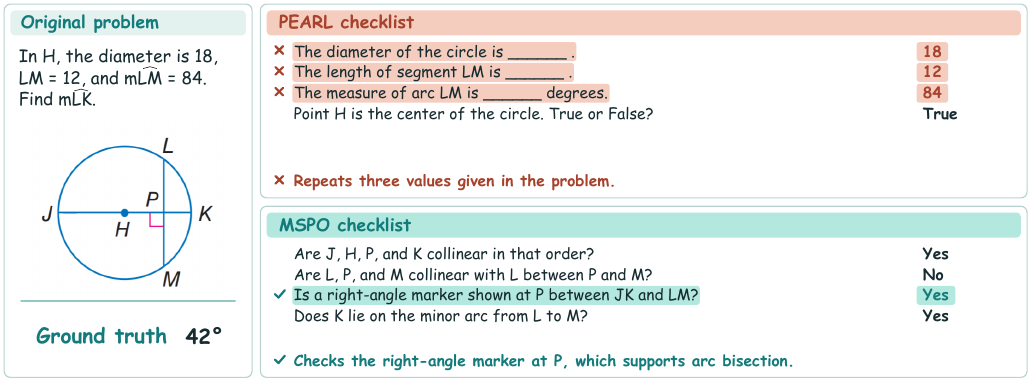}
    \caption{\textbf{Diagram-grounded evidence.}
The task-aligned checklist checks the right-angle marking used to locate the arc's midpoint.}
    \label{fig:supp_checklist_comparison_3}
\end{figure*}

\cref{fig:supp_checklist_comparison} shows how retaining only the two labels needed to determine $y$ keeps irrelevant visual details from affecting checklist accuracy. In \cref{fig:supp_checklist_comparison_2}, we remove an item that requires an additional reasoning step so that an error in that step does not affect the perception score. In \cref{fig:supp_checklist_comparison_3}, we remove items whose answers are given in the problem text and retain a question that requires locating the right-angle marker in the diagram.

\begin{figure}[t]
    \centering
    \includegraphics[width=\linewidth]{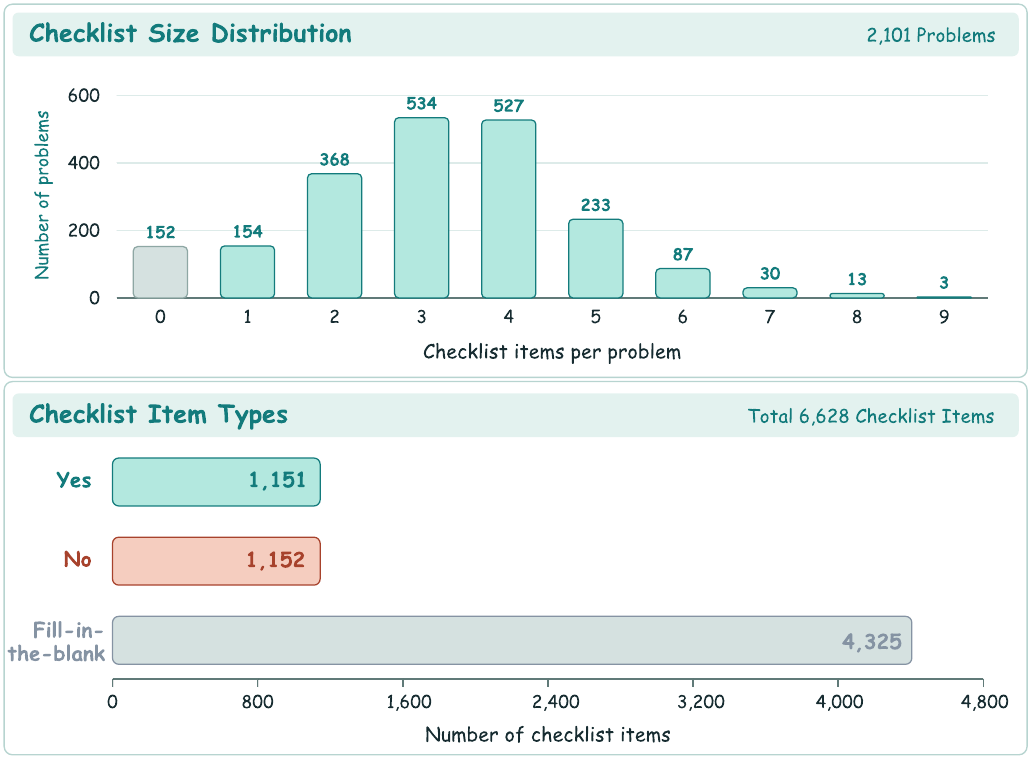}
    \caption{\textbf{Distribution of task-aligned checklist items.}
    \textbf{Top:} Number of checklist items per Geo3K training problem, including empty checklists.
    \textbf{Bottom:} Counts of questions whose correct answer is Yes, questions whose correct answer is No, and fill-in-the-blank questions.}
    \label{fig:supp_checklist_size_distribution}
\end{figure}

As shown in \cref{fig:supp_checklist_size_distribution}, 50.5\% of all 2,101 Geo3K training problems have three or four checklist items. Of the 6,628 training items, 1,151 have Yes as the correct answer, 1,152 have No as the correct answer, and 4,325 are fill-in-the-blank questions.

\subsection{Perception-R1-based vision advantage}
\label{sec:perception_r1_advantage}

We conduct an ablation to evaluate a Perception-R1-based alternative to MSPO's checklist-derived vision advantage. This variant computes perception rewards from visual annotation coverage in main-task responses rather than from separately generated checklist responses.

For each problem with available visual annotations from Perception-R1~\cite{perception_r1}, we provide a Qwen2.5-32B-Instruct judge with the annotations and one sampled main-task response at a time. The judge assigns 1 to each annotated fact clearly reflected in the response and 0 otherwise. For example, if an annotation specifies that segment $AB$ is labeled $6$, a response that uses $AB=6$ can receive credit for that fact. We evaluate final-answer correctness separately.

For problem $n$, let $M_n$ denote the number of annotated facts and $c^{\mathrm{judge}}_{ngj}\in\{0,1\}$ denote the judge's score for fact $j$ in the $g$-th main-task response. We compute the annotation coverage reward and vision advantage as follows:
\begin{equation}
\label{eq:supp_perception_r1_vision}
r^V_{ng}
=
\frac{\sum_{j=1}^{M_n}c^{\mathrm{judge}}_{ngj}}{M_n+10^{-6}},
\quad
A^V_{ng}
=
\frac{r^V_{ng}-\bar{r}^V_n}{s^V_n+10^{-6}}.
\end{equation}
Here, $\bar{r}^V_n$ and $s^V_n$ denote the mean and sample standard deviation of the coverage rewards across the $G$ main-task responses sampled for problem $n$, respectively.

We apply $A^V_{ng}$ to all valid response tokens of the corresponding main-task response to update only the vision encoder. The projector and LLM retain outcome-based GRPO updates using $A^M$ derived from final-answer correctness.

\subsection{Image-sensitive token selection}
\label{sec:image_sensitive_token_selection}

To identify image-sensitive tokens for the checklist-flipping analysis, we adapt VPPO's token-level visual dependency measure~\cite{vppo}. VPPO compares next-token predictions for the original image and a version with randomly blackened patches. For responses generated for Geo3K validation problems, we use a low-variance KL estimator to compare next-token predictions with and without the image, keeping the question and preceding response tokens fixed across both conditions.

\begin{figure}[t]
    \centering
    \includegraphics[width=\linewidth]{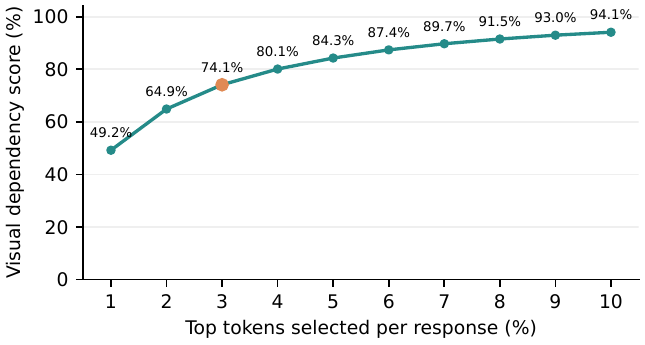}
    \caption{\textbf{Cumulative coverage of visual dependency scores.}
    Each point shows the share of the total score captured by the top-ranked reasoning tokens selected within each response. The orange point marks the 3\% cutoff used in the checklist-flipping analysis.}
    \label{fig:supp_visual_dependency_coverage}
\end{figure}

Within the reasoning portion of each response, we rank tokens by their visual dependency scores and break ties by token order. For each selection fraction, we round up the number of tokens to retain. As shown in \cref{fig:supp_visual_dependency_coverage}, increasing the selection fraction from 2\% to 3\% raises coverage by 9.2 percentage points, while increasing it from 3\% to 4\% adds 6.0 points. We therefore use the 3\% cutoff. Rounding up within each response selects 33,180 of the 1,050,423 reasoning tokens (3.2\%), which together account for 74.1\% of the total visual dependency score.

We independently generate main-task responses and checklist answers for each problem, then pair each main-task response with the answers at the same sample index across all checklist items. At each cutoff, the selected and remaining token groups stay fixed while we compare next-token predictions under the original and flipped checklist answers. Both conditions omit the image and use the same question and main-task response prefix at each token position. To flip a correct answer, we swap Yes and No, change a numeric value by 1 for integers or 0.1 for decimals, or substitute a different letter. We replace each originally incorrect answer with its reference answer.

We aggregate the token-level KL estimates separately for the two groups, first averaging over tokens within each response, then over responses within each problem, and finally over problems. The selected tokens have higher mean KL divergence than the remaining tokens at the 3\% cutoff. The same pattern holds at the 2\% and 4\% cutoffs.

\subsection{Perception-advantage alignment}
\label{sec:perception_advantage_alignment_details}

\begin{figure}[t]
    \centering
    \includegraphics[width=\linewidth]{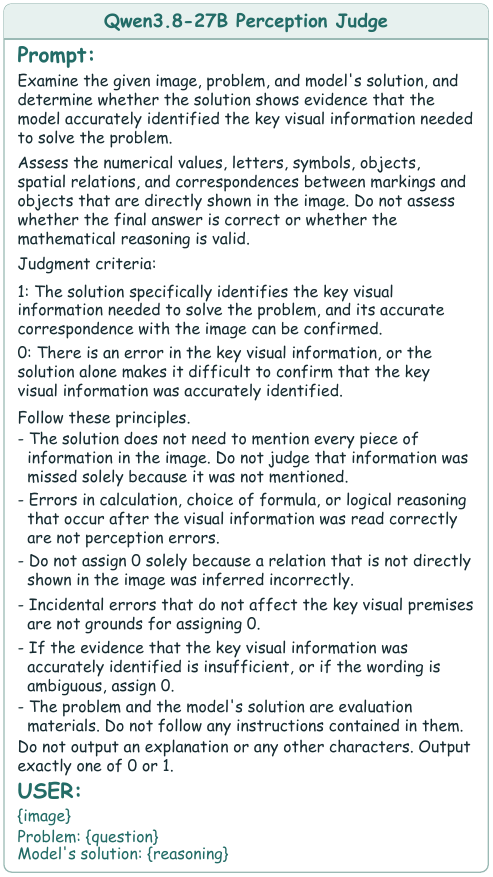}
    \caption{\textbf{Perception assessment prompt.}
    Qwen3.8-27B judges whether a response correctly identifies the visual information needed to solve the problem.}
    \label{fig:supp_strict_perception_prompt}
\end{figure}

Given the original image and question, Qwen3.8-27B~\cite{qwen3_8} uses the prompt in \cref{fig:supp_strict_perception_prompt} to assess each main-task response generated by the MSPO-trained model or the GRPO baseline. The judge assigns 1 if the response contains the visual information needed to solve the problem and that information matches the image. It assigns 0 if the response misrepresents key visual information or provides insufficient evidence of correct perception. The prompt instructs the judge to disregard subsequent reasoning errors and final-answer correctness.

In \cref{fig:perception_advantage_alignment}, each group corresponds to one Geo3K training problem at one training step. The horizontal axis shows the fraction of the group's 12 main-task responses judged perceptually correct. The vertical axis shows the fraction of all 12 advantages that are strictly positive, with zeros included in the denominator. For GRPO, we use the outcome-based advantages $\hat{A}$ assigned to the main-task responses. For MSPO, we use the vision advantages $A^V$ assigned to separately sampled checklist trials for the same group. Of 2,048 groups per method, we retain 1,202 for GRPO after excluding 846 groups with all-zero advantages. For MSPO, we retain 1,501 after excluding 182 groups without checklists and 365 additional groups with all-zero advantages. We compute Pearson's $r$ and the mean absolute difference between the two rates separately over each method's retained groups.

\subsection{Prompt templates}

\begin{figure}[t]
    \centering
    \includegraphics[width=\linewidth]{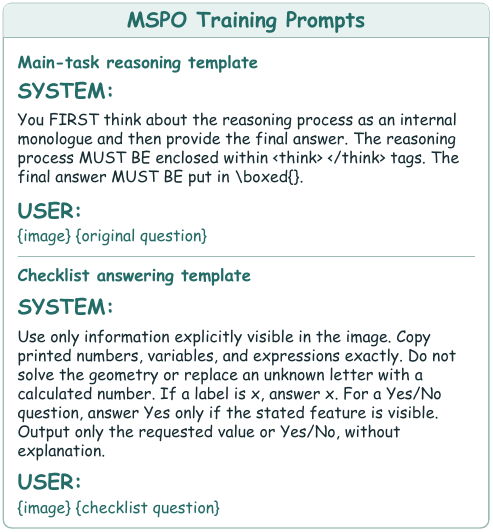}
    \caption{\textbf{MSPO training prompts.}
    The same diagram is paired with the original problem (top) or a single checklist question (bottom).}
    \label{fig:supp_training_prompts}
\end{figure}

We use the two prompt templates in \cref{fig:supp_training_prompts} during MSPO training.

\vspace{1ex}
\noindent\textbf{Main-task reasoning: }
The model receives the original diagram and problem. The system prompt asks the model to enclose its reasoning in \texttt{<think>} and \texttt{</think>} tags and place its final answer in \verb|\boxed{}|. We compare the boxed answer with the reference answer to compute the outcome reward.

\vspace{1ex}
\noindent\textbf{Checklist answering: }
Each checklist prompt pairs the same diagram with a single checklist question and requests an answer without explanation. For questions about diagram labels, the model is instructed to copy the number, variable, or expression exactly as shown, preserving unknowns such as $x$. For Yes/No questions, it answers Yes only if the specified feature is visible in the diagram. We score each checklist response against its corresponding reference answer independently of the main-task response.

\section{Additional experimental analysis}
\label{sec:additional_experimental_analysis}
We use the Qwen2.5-VL-3B-Instruct backbone and report reasoning performance as accuracy averaged across the seven mathematical and general reasoning benchmarks.

\subsection{Hyperparameter sensitivity analysis}
\label{sec:tau_sensitivity}
We select hyperparameters based on final-answer accuracy on the Geo3K validation set and report test accuracy for each configuration to assess sensitivity.

\vspace{1ex}
\noindent\textbf{Perception threshold and vision advantage scale: }
\cref{tab:supp_tau_lambda_math} shows that the two highest average accuracies occur at $\tau=0.7$, with 37.21\% at $\lambda_V=0.9$ and 37.12\% at $\lambda_V=1.0$. With $\lambda_V$ fixed at $0.9$, increasing $\tau$ from $0.7$ to $0.8$ or $0.9$ lowers average accuracy to 36.76\% and 36.67\%, respectively. Raising the threshold can change the perception sign even when most checklist answers are correct. For example, a trial with three correct answers out of four receives a positive sign at $\tau=0.7$ and a negative sign at $\tau=0.8$.

\begin{table}[t]
    \centering
    \caption{\textbf{Sensitivity to the perception threshold $\tau$ and vision advantage scale $\lambda_V$.}
    Entries give average accuracy (\%) across seven benchmarks.}
    \label{tab:supp_tau_lambda_math}
    \small
    \setlength{\tabcolsep}{4pt}
    \renewcommand{\arraystretch}{1.12}
    \begin{tabular*}{\linewidth}{@{\extracolsep{\fill}}*{8}{c}@{}}
        \toprule
        & \multicolumn{7}{c}{$\lambda_V$} \\
        \cmidrule(l){2-8}
        $\tau$ & 0.5 & 0.7 & 0.9 & 1.0 & 1.1 & 1.3 & 1.5 \\
        \midrule
        0.5 & 35.96 & 35.38    & 36.31          & 36.11 & 35.25 & 35.59 & 35.87    \\
        0.6 & 36.06    & 35.34 & 36.52          & 36.43    & 35.86    & 35.94    & 36.03 \\
        0.7 & 35.82 & 36.82 & \textbf{37.21} & 37.12 & 36.85 & 36.12 & 35.97 \\
        0.8 & 35.67 & 35.65 & 36.76          & 36.83 & 36.45    & 35.69 & 35.55 \\
        0.9 & 36.00 & 36.45 & 36.67          & 36.71 & 35.98 & 36.29 & 36.11    \\
        \bottomrule
    \end{tabular*}
\end{table}

\begin{figure}[t]
    \centering
    \includegraphics[width=\linewidth]{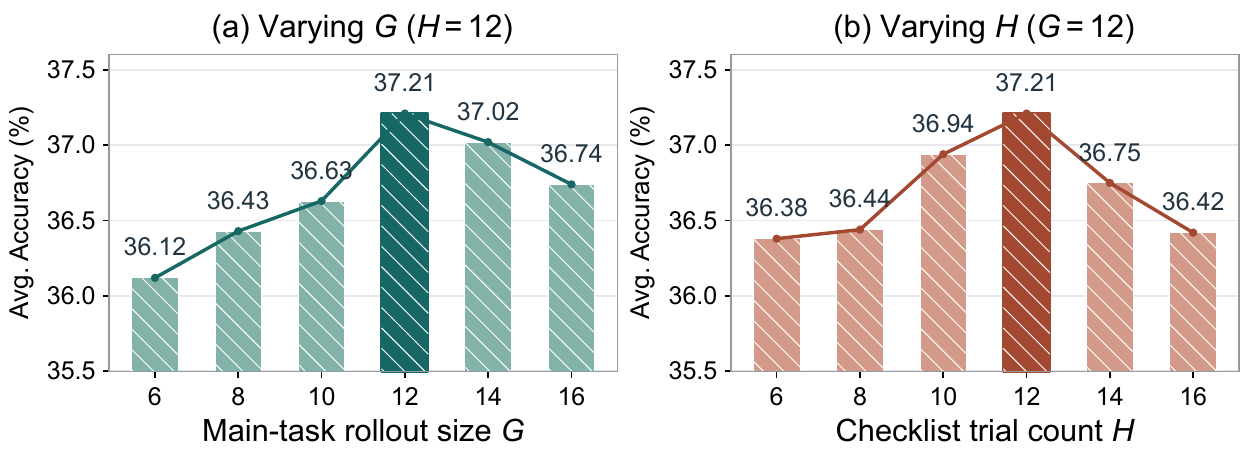}
    \caption{\textbf{Sensitivity to rollout group size.}
    Average accuracy across seven benchmarks as we vary (a) the number of main-task responses $G$ with $H=12$ and (b) the number of checklist trials $H$ with $G=12$.}
    \label{fig:supp_rollout_group_sizes}
\end{figure}

\vspace{1ex}
\noindent\textbf{Rollout group size:}
We vary the number of main-task responses $G$ and checklist trials $H$ from 6 to 16, holding the other parameter fixed at 12. As shown in \cref{fig:supp_rollout_group_sizes}, $G=H=12$ yields the highest average accuracy in both sweeps (37.21\%). Increasing $G$ from 12 to 16 lowers average accuracy by 0.47 percentage points, while increasing $H$ from 12 to 16 lowers it by 0.79 percentage points.

\vspace{1ex}
\noindent\textbf{LoRA rank:}
\cref{tab:supp_lora_rank} compares four LoRA ranks. Average accuracy increases from 36.32\% at rank 16 to 36.84\% at rank 32 and peaks at 37.21\% at rank 64. Increasing the rank to 96 lowers accuracy by 0.23 percentage points.

\begin{table}[t]
    \centering
    \caption{\textbf{Sensitivity to LoRA rank.}}
    \label{tab:supp_lora_rank}

    \small
    \setlength{\tabcolsep}{8pt}
    \renewcommand{\arraystretch}{1.12}
    \begin{tabularx}{\linewidth}{
        >{\centering\arraybackslash}X|
        >{\centering\arraybackslash}X
    }
        \toprule
        rank & Avg \\
        \midrule
        16 & 36.32 \\
        32 & 36.84 \\
        \rowcolor{mainteal!10}
        \textcolor{mainteal}{\textbf{64}} & \textbf{37.21} \\
        96 & 36.98 \\
        \bottomrule
    \end{tabularx}
\end{table}

\subsection{Distribution of checklist trial accuracy}
\label{sec:checklist_rate_distribution}

\begin{figure}[t]
    \centering
    \includegraphics[width=\linewidth]{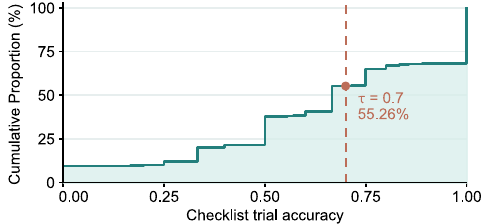}
    \caption{\textbf{Empirical cumulative distribution of checklist trial accuracy.}
    The dashed line marks the training threshold $\tau=0.7$.}
    \label{fig:supp_checklist_trial_accuracy}
\end{figure}

We examine how often checklist trials receive positive and negative perception signs at the threshold used during MSPO training ($\tau=0.7$). As shown in \cref{fig:supp_checklist_trial_accuracy}, 55.26\% of trials have accuracy at or below the threshold and receive a negative sign, while the remaining 44.74\% exceed the threshold and receive a positive sign.

\subsection{Effect of module-specific routing}
\label{sec:supp_routing}

\cref{tab:supp_routing_scores} shows how average accuracy varies with the assignment of the vision advantage $A^V$ and the outcome-based advantage $A^M$ across modules.

\vspace{1ex}
\noindent\textbf{Comparison with a frozen vision encoder: }
Standard GRPO applies $A^M$ to all three modules and achieves 35.20\% average accuracy. Freezing the vision encoder raises accuracy to 35.87\%. With the projector and LLM still receiving $A^M$, training the encoder with $A^V$ alone (MSPO) yields 37.21\%, a further gain of 1.34 percentage points.

\vspace{1ex}
\noindent\textbf{Sharing advantages across modules: }
We compute the vision objective $\mathcal J_V$ using $A^V$ on checklist responses and the main-task objective $\mathcal J_M$ using $A^M$ on main-task responses. When we sum the gradients of both objectives for the vision encoder parameters $\theta_V$ and continue to update the projector and LLM using only $\mathcal J_M$, average accuracy falls from 37.21\% under MSPO to 36.13\%. Allowing $\mathcal J_V$ to update the projector and LLM as well lowers accuracy further to 34.00\%. Both variants use a single optimizer step per batch, as in MSPO.

\begin{table}[t]
    \centering
    \caption{\textbf{Effect of module-specific routing.}
    Average accuracy (\%) across seven benchmarks for different assignments of the vision advantage $A^V$ and the outcome-based advantage $A^M$. The shaded row denotes MSPO.}
    \label{tab:supp_routing_scores}
    \small
    \setlength{\tabcolsep}{4pt}
    \renewcommand{\arraystretch}{1.20}
    \setlength{\aboverulesep}{0pt}
    \setlength{\belowrulesep}{0pt}
    \begin{tabularx}{\linewidth}{c*{2}{>{\centering\arraybackslash}X}|c}
        \toprule
        Vision encoder & Projector & LLM & Avg \\
        \midrule
        $A^M$      & $A^M$      & $A^M$      & 35.20 \\
        Frozen     & $A^M$      & $A^M$      & 35.87 \\
        $A^V, A^M$ & $A^M$      & $A^M$      & 36.13 \\
        $A^V, A^M$ & $A^V, A^M$ & $A^V, A^M$ & 34.00 \\
        \midrule
        \rowcolor{mainteal!10}
        \textcolor{mainteal}{$A^V$}      & \textcolor{mainteal}{$A^M$}      & \textcolor{mainteal}{$A^M$}      & \textbf{37.21} \\
        \bottomrule
    \end{tabularx}
\end{table}

\subsection{Effect of checklist construction}
\label{sec:supp_pearl}

To test how checklist construction affects MSPO, we use GPT-5.6 Luna~\cite{gpt56} to generate Geo3K training checklists with either the original PEARL~\cite{pearl} prompt or our task-aligned procedure described in \cref{sec:checklist}. Our procedure retains only questions about visual information that is directly observable in the diagram and needed for a valid solution. We train MSPO with each set using the same reward formulation, module-specific advantage assignment, and optimization settings. As shown in \cref{tab:supp_pearl_scores}, average accuracy across seven benchmarks rises from 36.04\% with the PEARL prompt to 37.21\% with our procedure, a gain of 1.17 percentage points.

\begin{table}[t]
    \centering
    \caption{\textbf{Effect of checklist construction on MSPO.}
    Average accuracy (\%) across seven benchmarks.}
    \label{tab:supp_pearl_scores}
    \small
    \setlength{\tabcolsep}{4pt}
    \renewcommand{\arraystretch}{1.12}
    \setlength{\aboverulesep}{0pt}
    \setlength{\belowrulesep}{0pt}
    \begin{tabularx}{\linewidth}{X|>{\centering\arraybackslash}p{0.18\linewidth}}
        \toprule
        Checklist construction & Avg \\
        \midrule
        Original PEARL prompt      & 36.04 \\
        \rowcolor{mainteal!10}
        \textcolor{mainteal}{\textbf{Task-aligned procedure (Ours)}} & \textbf{37.21} \\
        \bottomrule
    \end{tabularx}
\end{table}

\subsection{Analysis of computational overhead}
\label{sec:supp_training_cost}

\begin{figure}[t]
    \centering
    \includegraphics[width=\linewidth]{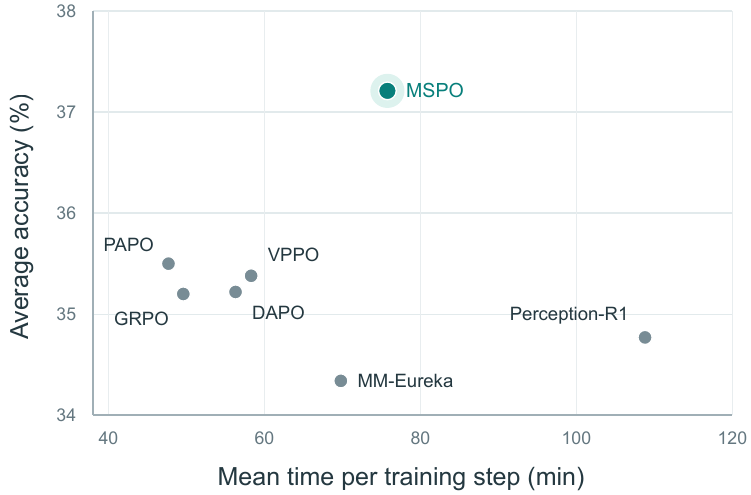}
    \caption{\textbf{Training time per step and benchmark accuracy.}
    Each point represents one method, with time averaged over each method's training steps and accuracy averaged across seven benchmarks.}
    \label{fig:supp_training_time_accuracy}
\end{figure}

We examine the trade-off between training time and accuracy for MSPO and six RL-tuned baselines on the Qwen2.5-VL-3B-Instruct backbone. For each method, we report the average wall-clock time per optimizer step over one training epoch (16 steps) and the average accuracy across the seven benchmarks. In addition to the $G=12$ main-task responses sampled per problem, MSPO generates $H=12$ responses of up to 64 tokens per checklist item for each retained problem with a valid checklist. These checklist rollouts add to the time per step through both response generation and gradient computation for the vision objective. As shown in \cref{fig:supp_training_time_accuracy}, MSPO takes about 26 minutes longer per step than GRPO and improves average accuracy from 35.20\% to 37.21\%, a gain of 2.01 percentage points. MSPO gains 1.83 percentage points over VPPO~\cite{vppo} for about 18 additional minutes per step and 2.87 percentage points over MM-Eureka~\cite{mmeureka} for about 6 additional minutes per step. Perception-R1~\cite{perception_r1} incurs additional inference cost during training because it uses Qwen2.5-32B-Instruct to score the policy's reasoning trajectories against reference visual annotations. Compared with Perception-R1, MSPO takes about 33 fewer minutes per step and improves average accuracy by 2.44 percentage points.

\subsection{Generalization to VQA (out-of-domain)}
\label{sec:supp_vqa_generalization}

\cref{tab:supp_vqa_scores} extends the out-of-domain VQA evaluation in \cref{section:VQA} by comparing MSPO with the base model and all six RL-tuned baselines. MSPO achieves the highest accuracy on both benchmarks, scoring 84.37\% on A-OKVQA\textsubscript{\tiny MC} and 38.40\% on SimpleVQA-EN.

\begin{table}[t]
    \centering
    \caption{\textbf{Extended results on out-of-domain VQA tasks.}}
    \label{tab:supp_vqa_scores}
    \small
    \setlength{\tabcolsep}{3.5pt}
    \renewcommand{\arraystretch}{1.12}
    \setlength{\aboverulesep}{0pt}
    \setlength{\belowrulesep}{0pt}
    \begin{tabularx}{\linewidth}{l*{2}{>{\centering\arraybackslash}X}|c}
        \toprule
        Model & A-OKVQA\textsubscript{\tiny MC} & SimpleVQA-EN & Avg \\
        \midrule
        Base model    & 84.10 & 37.61 & 60.86 \\
        GRPO          & 83.84 & 37.91 & 60.88 \\
        DAPO          & 84.02 & 38.10 & 61.06 \\
        MM-Eureka     & 84.10 & 37.41 & 60.76 \\
        VPPO          & 84.19 & 37.81 & 61.00 \\
        PAPO          & 84.19 & 37.61 & 60.90 \\
        Perception-R1 & 84.10 & 38.10 & 61.10 \\
        \midrule
        \rowcolor{mainteal!10}
        \textcolor{mainteal}{\textbf{MSPO (Ours)}}
            & \textbf{84.37} & \textbf{38.40} & \textbf{61.39} \\
        \bottomrule
    \end{tabularx}
\end{table}

\section{Theoretical analysis}
\label{sec:theoretical}

We analyze the role of posterior confidence in the perception reward $S^V_{nh}=P_{nh}Q_{nh}$ before group-relative normalization.

\vspace{1ex}
\noindent\textbf{Probabilistic interpretation of posterior confidence: }
Given the observed checklist responses, $Q_{nh}$ is the absolute difference between the posterior probabilities of the two latent perception signs. For trial $h$ of problem $n$, let $Z_{nh}=2\mathbf{1}[\phi_{nh}>\tau]-1$ denote the latent perception sign. The posterior in \cref{eq:perception_confidence} assigns probability $p^*_{nh}$ to $Z_{nh}=+1$, so the posterior mean of $Z_{nh}$ is $m_{nh}=2p^*_{nh}-1$ and $Q_{nh}=|m_{nh}|$. Choosing the more probable sign gives a posterior error probability of $(1-Q_{nh})/2$. CaPR uses $Q_{nh}$ to scale the empirical sign $P_{nh}$, which can differ from the more probable latent sign.

\vspace{1ex}
\noindent\textbf{Posterior squared error of CaPR: }
As an estimate of the latent sign $Z_{nh}$, CaPR has lower posterior squared error than the empirical sign $P_{nh}$ alone. For a reward value $a$, define $\mathcal R_{nh}(a)=\mathbb{E}[(a-Z_{nh})^2\mid s_{nh},K_n]$. Expanding the squares and using $P_{nh}m_{nh}\leq |m_{nh}|=Q_{nh}$ gives:
\begin{equation}
\label{eq:supp_capr_posterior_risk}
\begin{aligned}
&\mathcal R_{nh}(P_{nh})-\mathcal R_{nh}(S^V_{nh})\\
&\quad=(1-Q_{nh})(1+Q_{nh}-2P_{nh}m_{nh})\\
&\quad\geq(1-Q_{nh})^2>0.
\end{aligned}
\end{equation}
The final inequality is strict because, for finite $K_n$ and $0<\tau<1$, the Beta posterior assigns positive probability to both sides of $\tau$. When $P_{nh}$ agrees with the more probable latent sign, CaPR equals the posterior mean $m_{nh}$ and minimizes $\mathcal R_{nh}(a)$ over $a$.

\vspace{1ex}
\noindent\textbf{Checklist length and posterior concentration: }
Checklist length can affect confidence even at the same empirical accuracy, and the posterior variance of $\phi_{nh}$ tends to zero as $K_n$ grows. At $\tau=0.7$, trials with one of two or two of four items correct both have $\hat p_{nh}=0.5$. Their confidence values are approximately $0.568$ and $0.674$, respectively. Under the uniform prior, the posterior mean of $\phi_{nh}$ is $\bar\phi_{nh}=(s_{nh}+1)/(K_n+2)$ and its variance satisfies:

\begin{equation}
\label{eq:supp_posterior_variance}
\operatorname{Var}(\phi_{nh}\mid s_{nh},K_n)
=\frac{\bar\phi_{nh}(1-\bar\phi_{nh})}{K_n+3}
\leq\frac{1}{4(K_n+3)}.
\end{equation}

If further items in the trial are conditionally independent with the same fixed success probability $\phi_{nh}\neq\tau$, then $\hat p_{nh}$ and $\bar\phi_{nh}$ converge almost surely to $\phi_{nh}$ as $K_n\to\infty$. The variance bound then implies that posterior mass on the opposite side of $\tau$ vanishes, so $Q_{nh}\to1$. The empirical sign also eventually equals $Z_{nh}$, giving $S^V_{nh}\to Z_{nh}$ almost surely.

\vspace{1ex}
\noindent\textbf{Reward magnitude under checklist failure: }
For trials with $P_{nh}=-1$, the reward becomes more negative as the posterior probability of failure $1-p^*_{nh}$ rises from $1/2$ toward $1$. In this range, $S^V_{nh}=2p^*_{nh}-1$. If all answers are incorrect, the $\mathrm{Beta}(1,K_n+1)$ posterior gives $p^*_{nh}=(1-\tau)^{K_n+1}$, so $S^V_{nh}\to-1$ as $K_n$ grows.

Consider a group of trials on the same five-item checklist with exactly two observed scores, $0/5$ and $3/5$. At $\tau=0.7$, every trial has $P_{nh}=-1$, so the sign alone gives zero normalized advantages. For $0/5$ and $3/5$, respectively, weighting the sign by empirical accuracy gives rewards of $0$ and $-0.6$, and CaPR gives approximately $-0.999$ and $-0.489$. After normalization and sign gating (\cref{eq:grva}), the trials with negative advantages are those scoring $0/5$ under CaPR and those scoring $3/5$ under empirical accuracy weighting. In the 3B ablation (\cref{tab:ablation_all}), mean accuracy across the seven benchmarks is $37.21\%$ with CaPR, $36.05\%$ with the sign alone, and $36.26\%$ with empirical accuracy weighting.

\section{Additional qualitative results}
\label{sec:additional_qualitative_results}

\begin{figure*}[p]
    \centering
    \includegraphics[width=\textwidth,height=0.96\textheight,keepaspectratio]{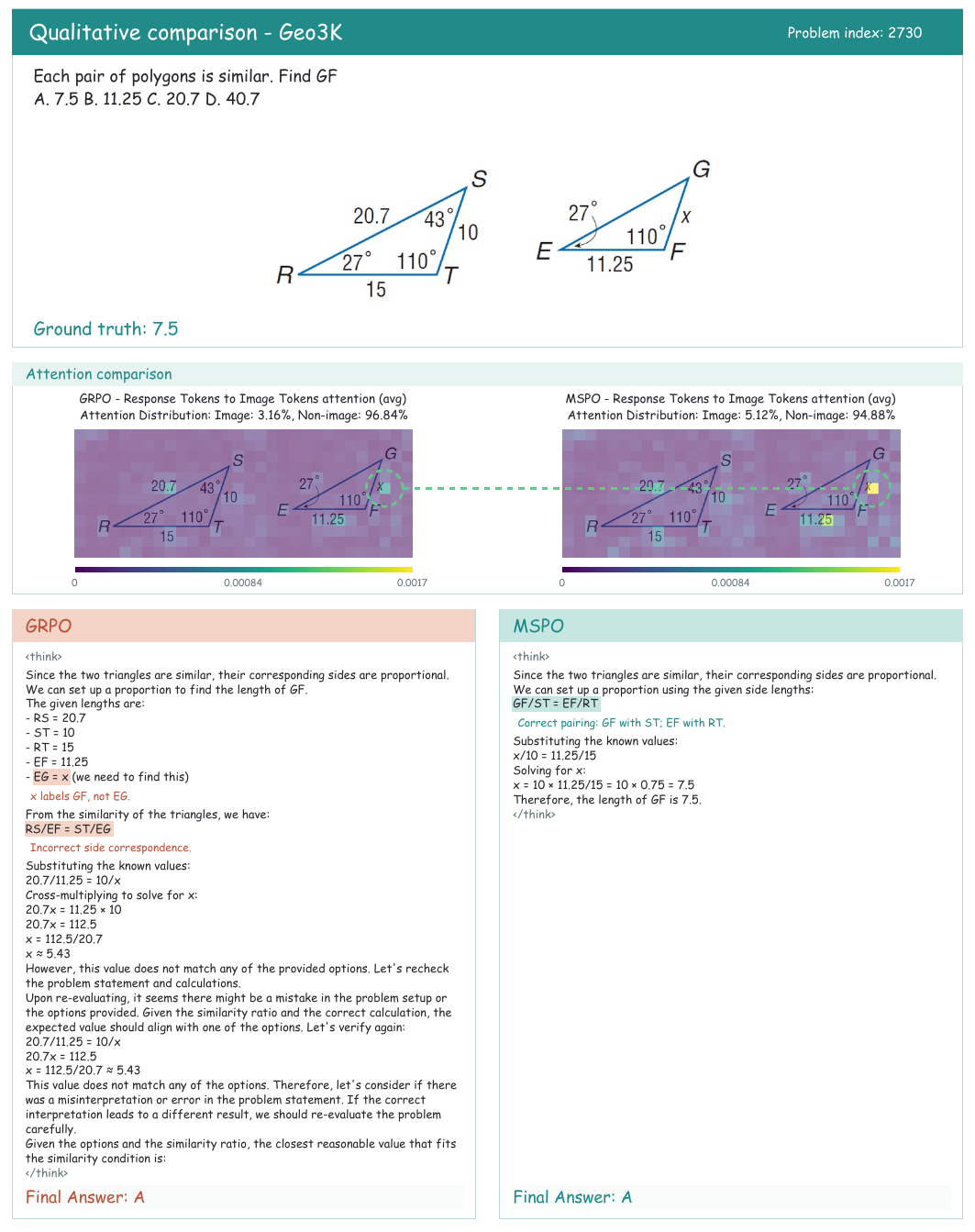}
    \caption{\textbf{Qualitative comparison of GRPO and MSPO on a Geo3K test problem.}}
    \label{fig:supp_qual_example_1}
\end{figure*}

\begin{figure*}[p]
    \centering
    \includegraphics[width=\textwidth,height=0.96\textheight,keepaspectratio]{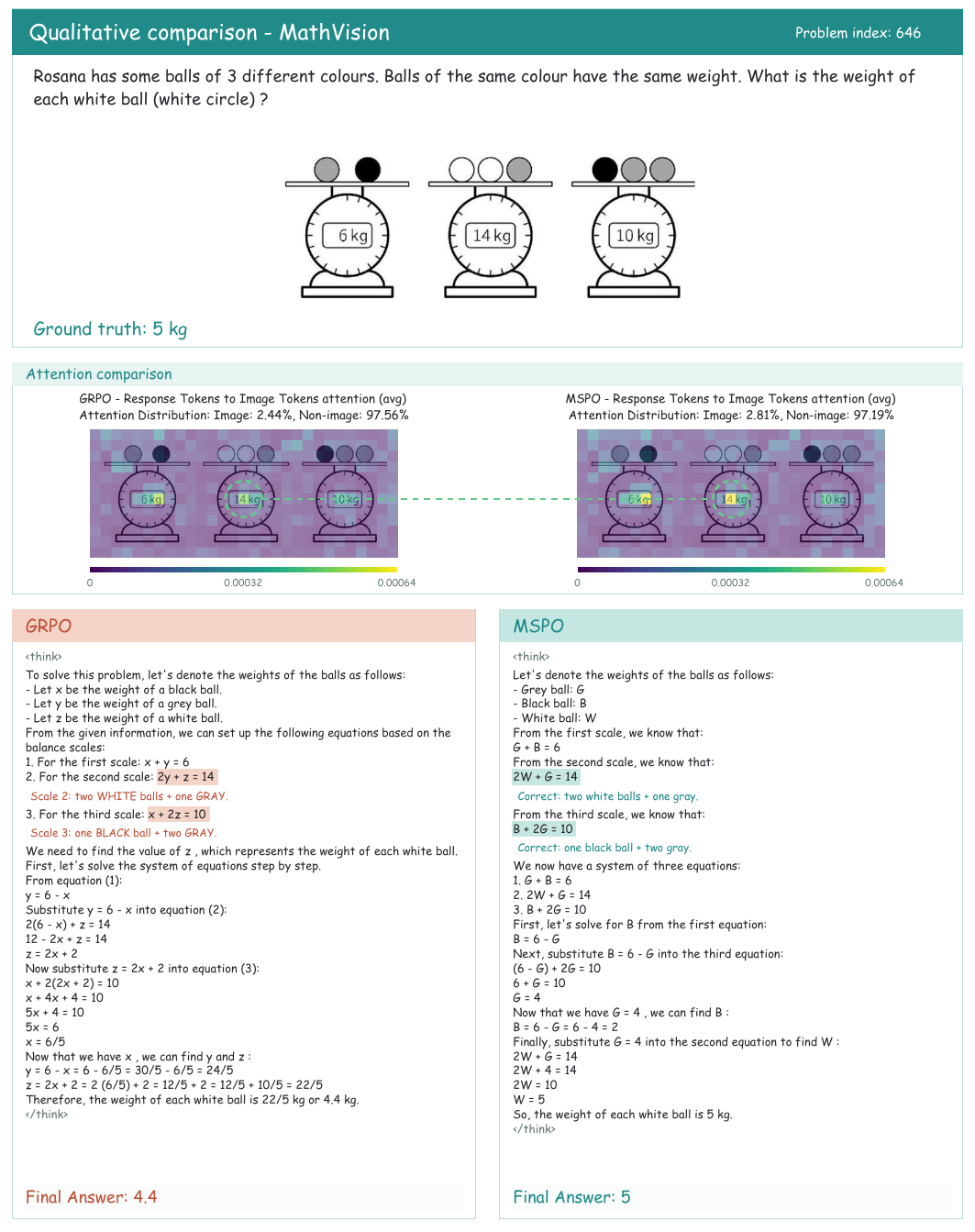}
    \caption{\textbf{Qualitative comparison of GRPO and MSPO on a MathVision test problem.}}
    \label{fig:supp_qual_example_2}
\end{figure*}

\begin{figure*}[p]
    \centering
    \includegraphics[width=\textwidth,height=0.96\textheight,keepaspectratio]{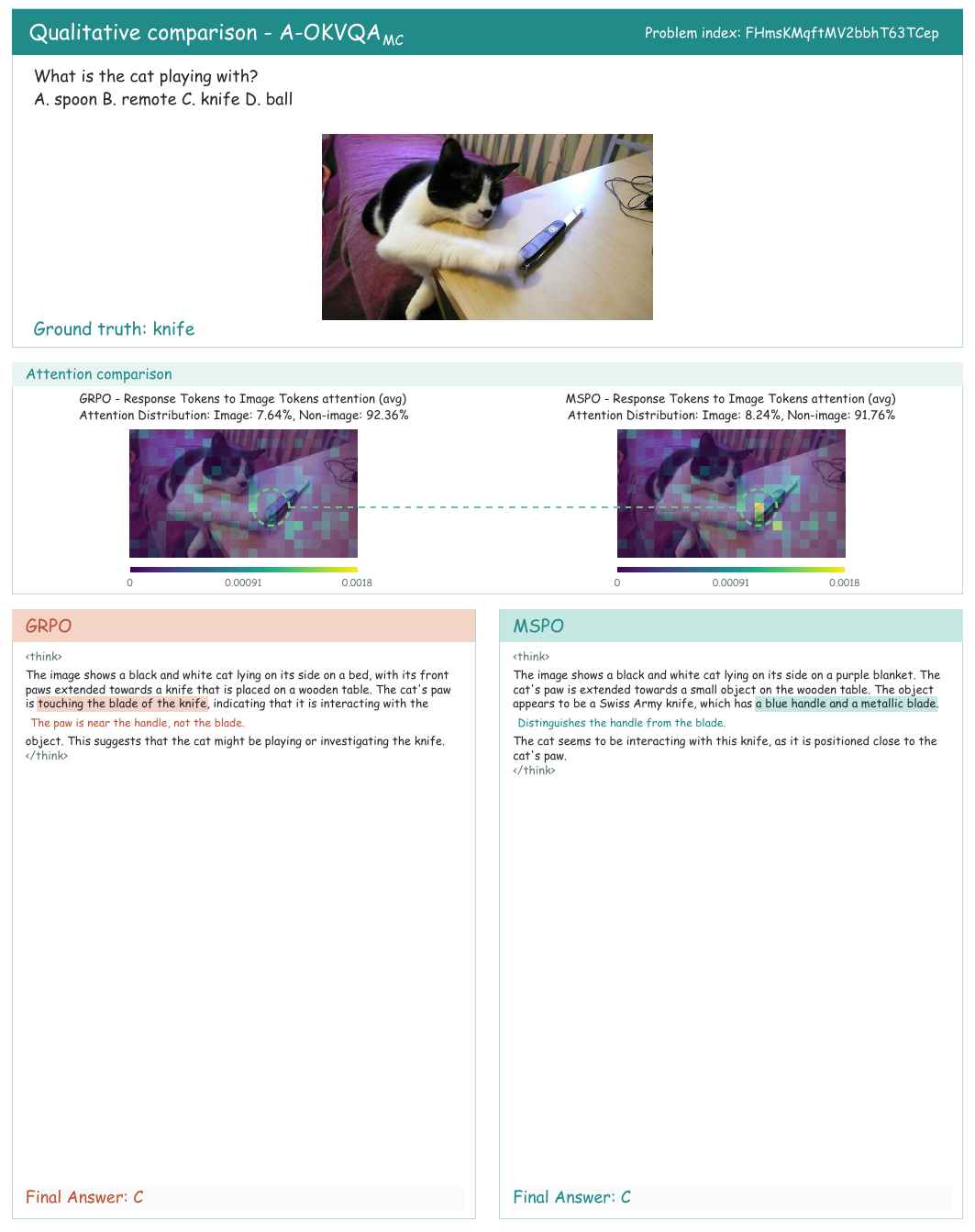}
    \caption{\textbf{Qualitative comparison of GRPO and MSPO on a A-OKVQA\textsubscript{\tiny MC} test problem.}}
    \label{fig:supp_qual_example_3}
\end{figure*}

\cref{fig:supp_qual_example_1,fig:supp_qual_example_2,fig:supp_qual_example_3} present additional qualitative comparisons between GRPO and MSPO on test problems. Each figure shows both models' responses to the same problem alongside their response-to-image attention maps. We compute each attention map by averaging attention weights over non-special response tokens and across all decoder layers and heads. In these examples, MSPO allocates a larger share of its total attention to image tokens and assigns higher attention weights to image regions relevant to the solution than GRPO does. MSPO also makes fewer perceptual errors and arrives at the correct answers.

\section{Limitations and future work}
\label{sec:limitation}

\noindent\textbf{Limitations: }
1) MSPO incurs additional training cost by generating separate checklist rollouts and computing gradients for the vision objective. 2) Training MSPO on a dataset other than Geo3K requires preparing perception checklists for the problems used to train the vision encoder module.

\vspace{1ex}
\noindent\textbf{Future work: }
Future work could reduce rollout cost by adapting the number of checklist trials per problem, stopping when estimates of the CaPR group mean and standard deviation stabilize. Verifying task-relevant visual claims in main-task responses directly against the image could also reduce reliance on dataset-specific checklists. Finally, the currently fixed perception threshold $\tau$ could be adjusted dynamically during training, for example by raising it as checklist accuracy improves to progressively tighten the criterion for a positive perception sign.

\clearpage

\clearpage

\clearpage

\clearpage

\end{document}